\documentclass[journal]{IEEEtran}

\usepackage{amsmath,amsfonts}
\usepackage{color}

\usepackage{array,multirow}
\usepackage{booktabs}

\usepackage[caption=false,font=normalsize,labelfont=sf,textfont=sf]{subfig}
\usepackage{textcomp}
\usepackage{stfloats}
\usepackage{url}
\usepackage{verbatim}
\usepackage{graphicx}
\usepackage{cite}
\usepackage[colorlinks,
            linkcolor=blue,   
            anchorcolor=blue, 
            citecolor=blue,      
            ]{hyperref}

\usepackage[table]{xcolor}
\usepackage{colortbl}       

\usepackage{graphicx}
\usepackage{caption}
\makeatletter
\newif\if@restonecol
\makeatother

\usepackage[linesnumbered,ruled,vlined]{algorithm2e}%[ruled,vlined]{
\usepackage{algpseudocode}
\usepackage{amsmath}
\graphicspath{{./Figs/}}

\begin{document}

\title{Robust Saliency-Aware Distillation for Few-shot Fine-grained Visual Recognition}

\author{Haiqi Liu, C. L. Philip Chen~\IEEEmembership{Fellow,~IEEE}, Xinrong Gong, Tong Zhang~\IEEEmembership{Senior Member,~IEEE}

        % <-this % stops a space
\thanks{
This work was funded in part by the National Natural Science Foundation of China grant under numbers 62076102, 62222603, and 92267203, in part by the National Key Research and Development Program of China under number 2019YFA0706200, in part by the STI2030-Major Projects grant from the Ministry of Science and Technology of the People’s Republic of China under number 2021ZD0200700, in part by the Key-Area Research and Development Program of Guangdong Province under number 2023B0303030001, in part by the Guangdong Natural Science Funds for Distinguished Young Scholar under number 2020B1515020041, and in part by the Program for Guangdong Introducing Innovative and Entrepreneurial Teams (2019ZT08X214). \textit{(Corresponding author: Tong Zhang)}
}

\thanks{The authors are with the School of Computer Science and Engineering, South China University of Technology, Guangzhou 510006, China, and is with the Brain and Affective Cognitive Research Center, Pazhou Lab, Guangzhou 510335, China. C. L. Philip Chen and Tong Zhang are also with the Engineering Research Center of the Ministry of Education on Health Intelligent Perception and Paralleled Digital-Human, Guangzhou 510006, China. (e-mail: tony@scut.edu.cn)}
}

\markboth{IEEE Transactions on Multimedia}%
{Shell \MakeLowercase{\textit{et al.}}: A Sample Article Using IEEEtran.cls for IEEE Journals}

\maketitle

\begin{abstract}
Recognizing novel sub-categories with scarce samples is an essential and challenging research topic in computer vision. Existing literature addresses this challenge by employing local-based representation approaches, which may not sufficiently facilitate meaningful object-specific semantic understanding, leading to a reliance on apparent background correlations. Moreover, they primarily rely on high-dimensional local descriptors to construct complex embedding space, potentially limiting the generalization. To address the above challenges, this article proposes a novel model, Robust Saliency-aware Distillation (RSaD), for few-shot fine-grained visual recognition. RSaD introduces additional saliency-aware supervision via saliency detection to guide the model toward focusing on the intrinsic discriminative regions. Specifically, RSaD utilizes the saliency detection model to emphasize the critical regions of each sub-category, providing additional object-specific information for fine-grained prediction. RSaD transfers such information with two symmetric branches in a mutual learning paradigm. Furthermore, RSaD exploits inter-regional relationships to enhance the informativeness of the representation and subsequently summarize the highlighted details into contextual embeddings to facilitate the effective transfer, enabling quick generalization to novel sub-categories. The proposed approach is empirically evaluated on three widely used benchmarks, demonstrating its superior performance. 

\end{abstract}


\section{Introduction}
\IEEEPARstart{F}{ine-grained} visual recognition (FGVR) is crucial and challenging research that aims to distinguish visually similar objects within a specific category, such as different species of birds or types of cars. With the significant advancements in computer vision\cite{lin2017feature, liufu2021reformative,dosovitskiy2021an}, extensive research\cite{ding2021ap,8907499,he2022transfg} has been conducted to improve the performance of FGVR. However, due to the considerable expenses entailed in fine-grained annotation and the relative scarcity of rare categories, the extensive applicability of such approaches is limited by the annotated data scarcity issue. Unlike machines, humans can distinguish subtle variations among previously unseen sub-categories, even when presented with one sample. Therefore, improving the generalizability of FGVR models in data-scarce scenarios is imperative, thereby reducing the disparities between humans and machines. Thus, this paper focuses on the few-shot fine-grained visual recognition (FS-FGVR) task, which involves utilizing scarce training examples per category (typically 1 or 5 samples) to recognize unseen similar-looking sub-categories. 
\begin{figure}[!t]
		\centering
        \includegraphics[scale=0.47]{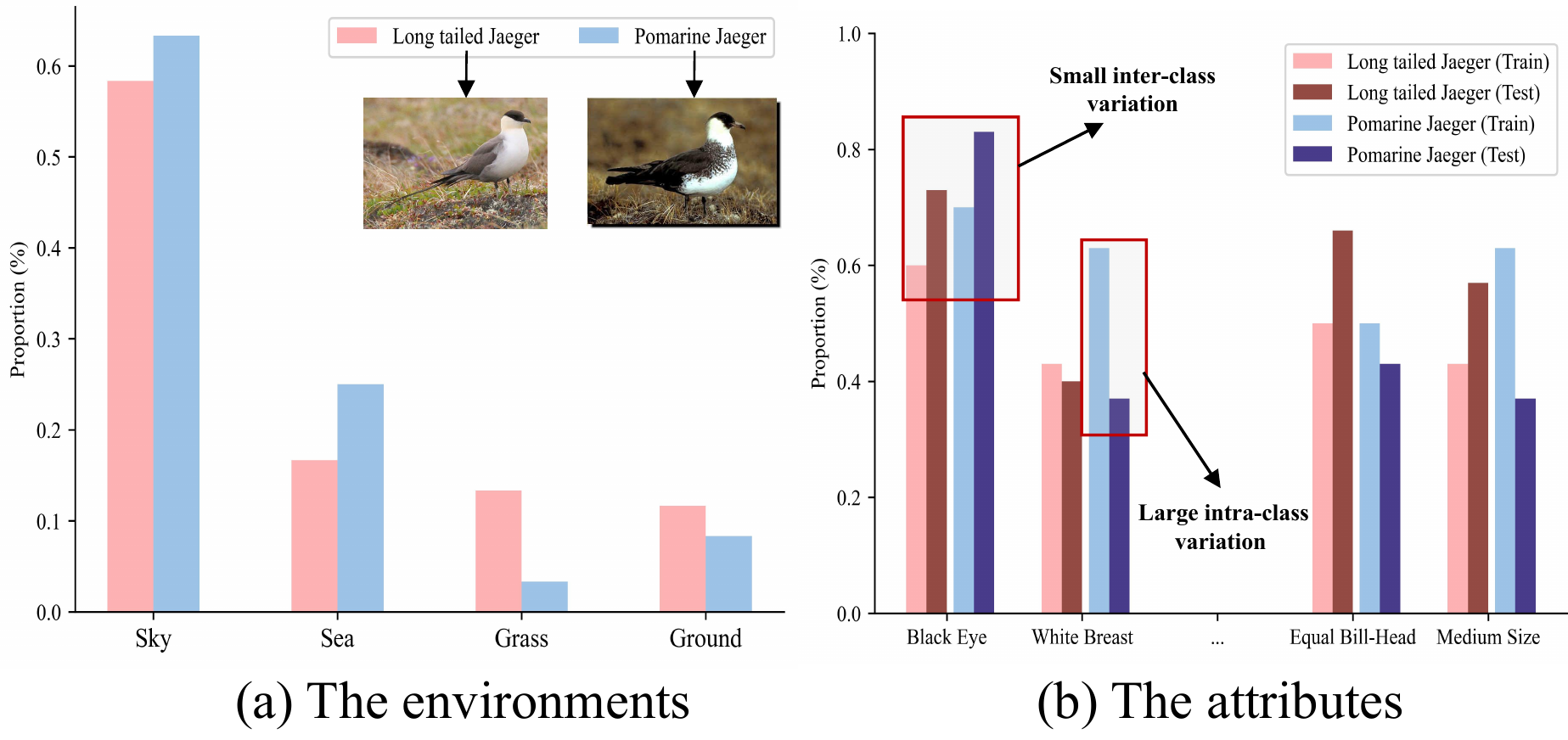}
		\caption{The statistical analysis of both external environments and intrinsic attributes across two sub-categories within the CUB-200-2011 dataset. The results reveal that 1) environments and specific attributes across different sub-categories may exhibit significant similarities; 2) within the same sub-category, identical attributes can vary considerably.}
		\label{fig:overview}
\end{figure}

The main challenge of FS-FGVR is quickly capturing key regions between similar-looking sub-categories. Existing works\cite{li2019revisiting,huang2021toan,wu2021object,wertheimer2021few} are mainly dedicated to solving this challenge in a local representation manner. In local representation-based approaches, the relationship between the support set (training data) and query set (testing data) is explored by the feature alignment~\cite{huang2021toan,wu2021object} or feature reconstruction~\cite{wertheimer2021few} to highlight semantic-related local features as the key regions. Despite the remarkable accomplishments in these approaches, there has been a lack of substantial focus on exploring the intrinsic structure (specifically, background and foreground) of fine-grained images, which are crucial factors for FGVR\cite{zhang2019few,wang2021fine,zhao2022boosting}. Fig.1(a) illustrates that environments across different sub-categories may exhibit notable similarities. The local-representation-based approach may inadequately promote object-relevant semantic understanding, thereby shifting attention towards apparent background correlations, which causes misclassification. Recent studies~\cite{zhang2019few,wang2021fine,zhao2021saliency,zhao2022boosting} have attempted to address this issue by integrating saliency detection models intended to redirect the model's attention to object-related features. However, these models generally apply saliency information for data or feature enhancement at the pixel or feature level. Such approaches present a considerable challenge in understanding high-dimensional saliency information with limited samples. The Deutsch–Norman late-selection filter model\cite{deutsch1963attention} in neuroscience indicates that humans will amplify meaningful information and weaken irrelevant information after processing it for meaning. Considering the previously stated observations, this paper designs a saliency-aware guidance strategy to encourage the model to focus on the object-relevant regions during prediction (low-dimensional space), thereby effectively reducing the negative impact of background clutter.

Furthermore, when confronted with novel fine-grained visual concept tasks, humans tend to identify noticeable distinctions and memorable attributes between them. It is rational that simple salient patterns can easily generalize to unseen sub-categories\cite{sorscher2022neural}. Previous approaches\cite{huang2021toan,wu2021object} have often relied on high-dimensional deep local descriptors to establish embedding spaces through Bilinear Pooling~\cite{kong2017low} and Relation Network~\cite{sung2018learning} techniques. Despite the fine-grained nature of these embedding spaces, their inherent space complexity poses challenges in terms of generalization and computational demands. Moreover, as depicted in Fig.1(b), it is evident that partial characters may not provide reliable clues for FS-FGVR. Some works\cite{yang2020d2n4,xu2021variational,xu2022dual} indicate that capturing the difference in holistic view can also effectively distinguish unseen similar-like visual concepts. Based on the above observations, this paper designed the representation highlight\&summarize module by learning a simplified contextual embedding space to ensure quick generalization towards novel sub-categories.

To address the above challenges, this paper proposes a Robust Saliency-aware Distillation model called RSaD to generate fine-grained intrinsic and transferable embedding for FS-FGVR. Specifically, RSaD consists of the Saliency-aware Guidance (SaG) strategy and Representation Highlight\&Summarize (RHS) module. The SaG is designed to exploit explicit intrinsic relationship of sub-categories via the distillation technique on the salient region probability distribution. The alignment of salient region probability distribution reduces the impact of background clutter, which provides more pure object information for fine-grained prediction. Subsequently, the RHS module is designed to highlight the object-relevant deep descriptors and squeeze them as the contextual embedding. It can effectively capture discriminative, simple patterns across sub-categories, facilitating rapid generalization to novel sub-categories while maintaining a low model complexity. Extensive experiments are conducted on widely used benchmarks to evaluate the proposed approach. Results indicate that the proposed model outperforms current state-of-the-art FS-FGVR approaches, confirming its effectiveness. To sum up, this work makes the following contributions:

\begin{enumerate}
    \item This article proposes a novel saliency-aware guidance strategy for FS-FGVR. This method first exploits object-specific information via the distillation technique on the salient region probability distribution, which provides more accurate clues for fine-grained prediction.
    \item This article proposes the RHS module, which involves learning a robust embedding space to facilitate rapid generalization towards novel sub-categories.
    \item Extensive experiments are conducted to verify the effectiveness of the proposed approach. The results demonstrate that the proposed approach achieves comparable performance with state-of-the-art approaches.
\end{enumerate}

The remainder of this article is as follows: Section~\ref{releted_work} provides a brief review of related work. Section~\ref{method} presents the details of the RSaD model. Section~\ref{exp_result} presents and analyzes comparative experiment results. Then, Section~\ref{discussion} discusses the impact of different modules proposed in this article. Finally, Section~\ref{conclusion} concludes with a summary and outlook.

\section{RELATED WORK}
\label{releted_work}
\subsection{Fined-Grained Visual Recognition}
Fine-grained Visual Recognition (FGVR) and fine-grained image retrieval~\cite{10197227} have emerged as critical components in fine-grained image analysis~\cite{9609630}. FGVR identifies visually similar objects within the same meta-category, typically at a sub-category or individual instance level. Early research~\cite{6248089,berg2013poof,gosselin2014revisiting} on FGVR was based on hand-crafted features such as attributes, SIFT\cite{lowe2004distinctive} and part annotations. However, the ability of such methods was somewhat limited due to the drawbacks of hand-crafted features, i.e., lack of information and high labeling effort\cite{chang2020devil}. Benefiting from the success of convolutional neural networks (CNNs), research on FGVR has shifted towards learning deep feature representations. For example, Lin \textit{et al.}~\cite{lin2015bilinear} introduced bilinear CNN models to generate rich deep features for FGVR, while Kong \textit{et al.}~\cite{Kong_2017_CVPR} designed a low-rank bilinear pooling strategy to reduce the computational effort required by high-dimensional bilinear features. Some previous works\cite{zhang2014part,zhang2016picking,yang2018learning,9591580} have leveraged object detection models such as FRN and R-CNN to select informative regions. These part-based approaches are practical in locating and analyzing fine-grained local features. However, these methods often incur significant computational costs and substantial data requirements, primarily attributable to the necessity of detailed analysis using auxiliary sub-networks or selective strategies~\cite{GUO2022107651}. More recently, several studies\cite{hu2021rams,zhang2022free,he2022transfg,sun2022sim} utilized vision transformer\cite{dosovitskiy2021an} to improve model capacity in FGVR. Regardless of the demonstrated successes, large-scale data is necessary for model training. Learning with limited data remains an open problem in FGVR. Thus, this paper focuses on the FS-FGVR task, which entails effectively leveraging limited training examples per sub-categories.

\subsection{Few-Shot Fine-Grained Visual Recognition}
Recently, research\cite{8752297, li2019revisiting} has emerged focusing on a more practical and challenging setting, termed FS-FGVR (Few-Shot Fine-Grained Visual Recognition). FS-FGVR aims to distinguish novel sub-categories with limited samples. Thanks to the success of few-shot learning~\cite{snell2017prototypical}, FS-FGVR research has made significant progress in recent years and can be broadly classified into two groups:

\textit{Global representation-based methods} aim to learn discriminative global features between sub-categories. Li \textit{et al.}~\cite{li2020bsnet} proposed the bi-similarity network, which considers various metric functions to find a better global feature space. Xu \textit{et al.}~\cite{xu2022dual} introduced a dual attention mechanism to provide a refined global embedding for recognition. Xu \textit{et al.}~\cite{xu2021variational} utilized $\beta$-TCVAE~\cite{chen2018isolating} to learn transferable intra-class variance to generate additional global features applicable to novel classes.

\textit{Local representation-based methods} focus on learning the discriminative parts of the whole image. They argue that global representations obtained by pooling lack critical spatial information in FS-FGVR. In local representation-based approaches, the relationship between the support set and query set is primarily leveraged through local descriptor alignment or local descriptor reconstruction to emphasize semantically related local features as discriminative features. Wei \textit{et al.}~\cite{8752297} proposed a novel exemplar-to-classifier mapping strategy that can learn a discriminative classifier within bilinear CNN features in a parameter-economic way. Li \textit{et al.}~\cite{li2019revisiting} replaced the traditional image-to-class measure with a local descriptor-based measure to capture differences at a finer spatial level. Wu \textit{et al.}~\cite{wu2021object} proposed an object-aware long-short-range spatial alignment strategy to align discriminative semantic parts between query and support sets. Zhang \textit{et al.}~\cite{zhang2022learning} explored cross-image object semantic relations to distinguish subtle feature differences. Wertheimer \textit{et al.}~\cite{wertheimer2021few} investigated the relationship between support-query pairs through local feature reconstruction.

Despite the promising results of the methods above, they either lack sufficient supervision signal for recognition or require complex structures to capture local relations. Moreover, there has been limited emphasis on systematically exploring the intrinsic structure, particularly the differentiation between background and foreground components, in FS-FGVR. Recently, several studies\cite{wang2021fine,zha2023boosting} have emerged that explicitly address the reduction of background clutter at the pixel level, leading to notable performance improvements. However, learning background noise filters at the pixel level, particularly with low model capacity, poses a significant challenge. Unlike the above work, this paper first leverages object-specific information in prediction by employing the distillation technique on the probability distribution of salient regions, thereby enhancing the precision of object information for fine-grained prediction.

\subsection{Saliency Detection}
Saliency detection is a computer vision technique that identifies regions of an image or video most visually significant to a human observer. Over the past few years, saliency detection has proven effective in various downstream tasks such as person re-identification~\cite{song2018mask} and visual tracking~\cite{9094040}. Several works have recently adopted saliency detection in few-shot learning research. Specifically, Zhang \textit{et al.}\cite{zhang2019few} employed a saliency detection model to separate foreground and background regions and generate additional samples by combining viable foreground-background. Wang \textit{et al.}\cite{wang2021fine} proposed a foreground object transformation strategy. They extracted the object's foreground by employing a saliency detection model and generating additional samples by the transformation learned in the foreground. Zhao \textit{et al.}~\cite{zhao2021saliency,zhao2022boosting} designed a complementary attention mechanism guided by saliency, which employs saliency detection signals to learn interpretable representation. In contrast to the above works, this article efficiently utilizes saliency detection signals from the viewpoint of knowledge distillation.

\subsection{Knowledge Distillation}
Knowledge distillation~\cite{hinton2015distilling} aims to transfer dark knowledge from a pre-trained large model to a smaller model, reducing its computational resource occupation. Existing works mainly include unidirectional knowledge distillation and bidirectional knowledge distillation. The former approach~\cite{Ahn_2019_CVPR,chen2021cross,zhao2022decoupled,chen2022knowledge} typically requires designing a solid network to serve as the teacher, generating soft labels to facilitate the student network's learning. However, obtaining strong teachers requires considerable effort, and negative transfer frequently arises due to the significant capacity gap between the teacher and student networks~\cite{stanton2021does}. Zhang \textit{et al.}~\cite{zhang2018deep} proposed a deep mutual learning paradigm in a bidirectional view to address these issues. Bidirectional knowledge distillation has drawn increasing attention in the community due to its simplicity and effectiveness~\cite{zhou2019omni,wei2020combating,Pang_2020_CVPR}. Despite the demonstrated achievements in knowledge distillation, previous works have primarily focused on scenarios with large sample sizes. In situations with limited data, the likelihood of negative transfer increases considerably due to high inductive bias. Ma \textit{et al.}~\cite{Ma_2021_ICCV} proposed partner-assisted learning, a two-stage learning scheme that transfers the good embedding space from the pre-trained partner to the few-shot learner. Zhou \textit{et al.}~\cite{zhou2021binocular} proposed a binocular mutual learning approach that benefits few-shot learners by incorporating global and local views. Ye \textit{et al.}~\cite{ye2022few} proposed the LastShot framework, which enables the few-shot learner to perform comparably to the pre-trained model. Differently, this article focuses on the distillation of saliency prior under the limited sample in a peer-teaching manner rather than relying on dark knowledge provided by the strong model.
\section{THE ADOPTED METHODOLOGY}
\label{method}
This section presents a robust saliency-aware distillation model for few-shot fine-grained visual recognition. The proposed model includes two core components: saliency-aware guidance and representation highlight\&summarize module. The details are explained as follows.
\begin{figure*}[t]
  \centering
   \includegraphics[width=0.8\textwidth]{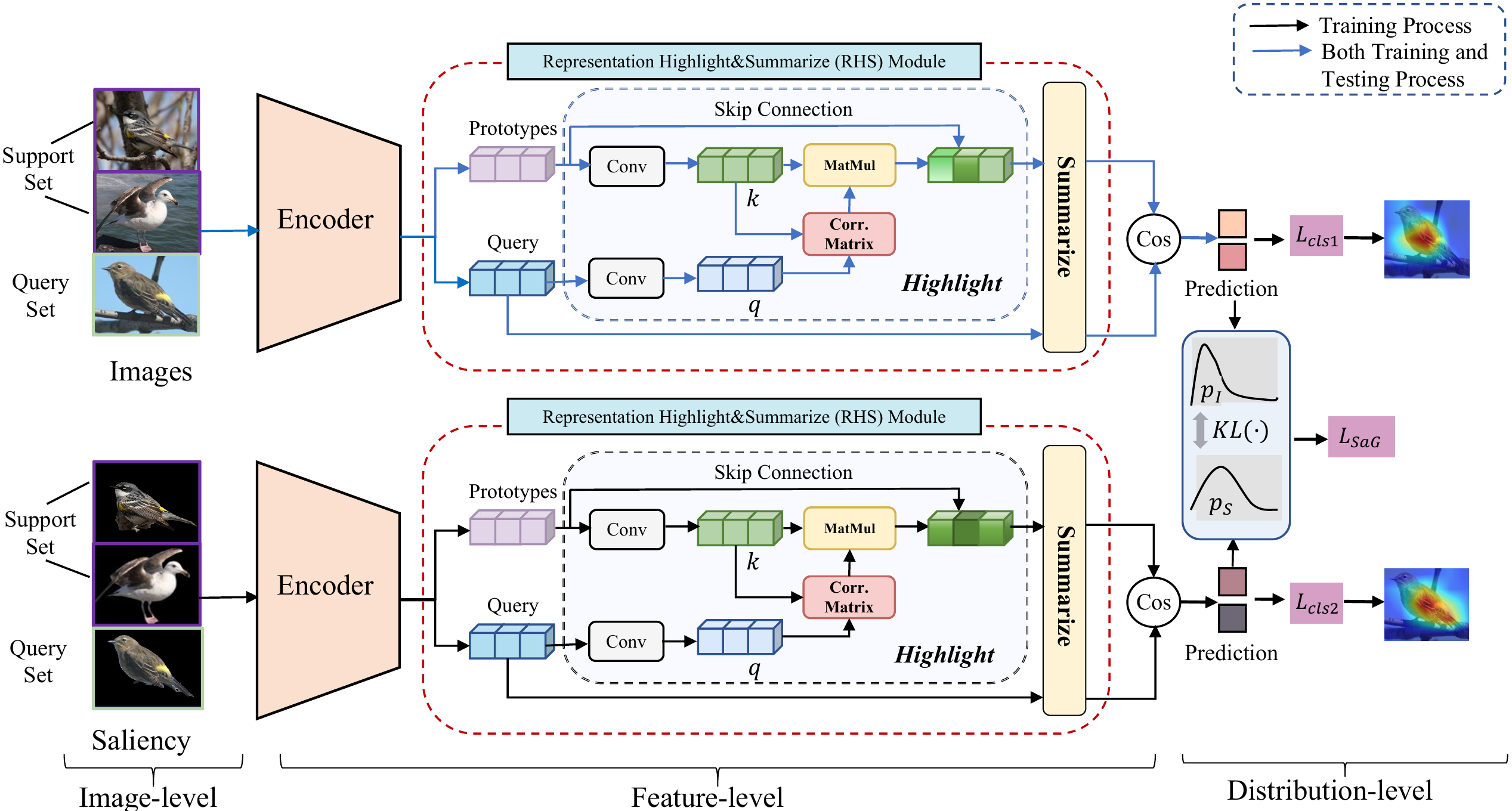}
   \caption{The framework of the proposed RSaD for few-shot fine-grained visual recognition. This framework consists of three hierarchical levels of operations. At the image level, the framework generates the saliency serving as input for the symmetric branch. At the feature level, it highlights crucial features while aggregating significant information. Finally, the two branches independently optimize the cross-entropy (CE) loss while simultaneously providing complementary signals at the distribution level via mutual learning.}
  \label{fig:sgml-net}
\end{figure*}

\subsection{Problem Definition}
FS-FGVR aims to learn general feature representation on large-scale sub-categories with annotations that can be generalized to unseen sub-categories. To simulate this scenario, the dataset $\mathcal{D}$ is typically divided into base set $\mathcal{D}_{\text {base}}$ and novel set $\mathcal{D}_{\text {novel}}$ by category, where $\mathcal{D}_{\text {base}} \cap \mathcal{D}_{\text {novel}}=\emptyset$. $\mathcal{D}_{\text {base}}$ represents the prior knowledge while $\mathcal{D}_{\text {novel}}$ represents the novel knowledge. For the typical FS-FGVR setup, models are trained (or tested) on $N$-way $K$-shot episodes sampled from ${\mathcal{D}_{\text{base}}}$ (or $\mathcal{D}_{\text{novel}}$). Each episode consists of a support set $\mathcal{D}^\mathcal{S}$ and a query set $\mathcal{D}^\mathcal{Q}$, sampled from the same $N$ class. The model updates its parameters using the support set $\mathcal{\mathcal{D}}^\mathcal{S}$ consisting of $K$ samples per class, where $K$ is set to $1$ or $5$ in few-shot setups. Moreover, the generalization performance of the updated model is evaluated using the query set $\mathcal{D}^\mathcal{Q}$. To be more specific, the general training objective can be expressed as 
\begin{equation}
L^*=\underset{L}{\arg \min } \sum_{\left(x,y\right) \in \mathcal{D}_{\text {base}}^{\mathcal{Q}}} \ell\left(L\left(x; \mathcal{D}_{\text {base }}^{\mathcal{S}}\right),y\right),
\label{objective}
\end{equation}
where $L$ refers to the few-shot learner, whereas $x$ and $y$ denote the samples and corresponding labels from  $\mathcal{D}_{\text {base }}^{\mathcal{S}}$, the support set in the base set. Additionally, $\ell$ represents the loss function, i.e., cross-entropy loss.

\subsection{Saliency-aware Guidance (SaG) for FS-FGVR}
Backgrounds typically have a negative impact on FS-FGVR. The backgrounds probably amplify the variance within a class and diminish the variance between classes, which misleads the learning of inherent discriminative feature embeddings. Since the saliency detection model effectively highlights image regions associated with human visual attention, it is natural to utilize these models to provide crucial supplementary supervision in the meta-training phase of few-shot learners. To this end, this article presents the SaG strategy, which introduces additional saliency-aware supervision to guide the model toward focusing on the intrinsic discriminative features. SaG consists of two steps: augmented saliency generation and saliency-aware knowledge transfer.
\subsubsection{Augmented Saliency Generation}
\begin{figure}[!t]
		\centering
        \includegraphics[scale=0.35]{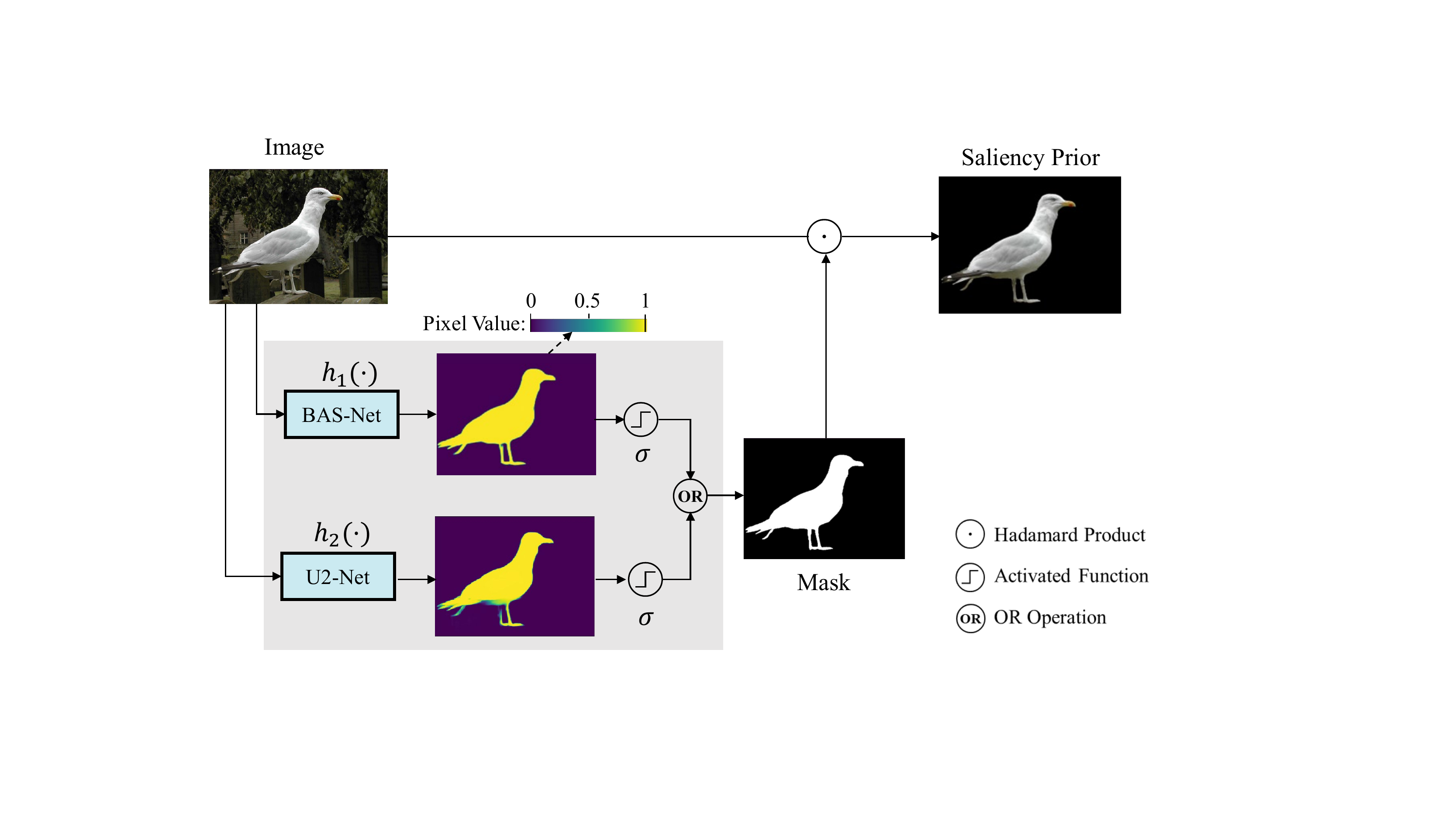}
		\caption{Augmented Saliency Generation. For the input image, this module generates multiple saliency maps. Next, binarization operations are performed on these maps, followed by an OR operation. Then, the priors are synthesized through the Hadamard product of the input image and mask.
}
		\label{fig:foa}
\end{figure}

To generate reliable saliency-aware supervision signals, it is crucial to ensure the quality of the generated saliency. Thus, this article employs multiple saliency detection models for image preprocessing and incorporates the resultant foreground as auxiliary data during model training. As shown in Fig.~\ref{fig:foa}, two saliency detection models, BAS-Net\cite{qin2019basnet} and U2-Net\cite{qin2020u2} are ensembled. Both are pre-trained on the DUTS-TR\cite{wang2017learning} dataset, which does not overlap with FS-FGVR benchmarks. The models  $h_1(\cdot)$ and $h_2(\cdot)$ process the given image $I$ and produce saliency maps with pixel values ranging from [0, 1], where a higher value indicates a more critical region. Next, binarization operations are performed on the two maps, followed by an OR operation to produce a binary mask. Then, the saliency of the image can be obtained by taking the Hadamard product of the image and the mask, as shown below:
\begin{equation}
    F_I=I \odot  \operatorname{Mask}(I),
\end{equation}
where $F_I$ represents the saliency prior. And $ \operatorname{Mask}(I)$ represents the mask of the image $I$ calculated by the following formula:
\begin{equation}
    \operatorname{Mask}(I)=\sigma\left(h_1(I)\right) \mid \sigma\left(h_2(I)\right),
\end{equation}
where the binary operation represented by $ \mid$ denotes logical OR, while the binarization process implemented by the activation function is denoted by $\sigma$. The activation function $\sigma$ is defined as follows:
\begin{equation}
    \sigma\left(h(I)_{i j}\right)=\left\{\begin{array}{l}
1, \text { if } h(I)_{i j} \geq t \\
0, \text { otherwise }
\end{array}\right.,
\end{equation}
where the value of the saliency map matrix in row $i$ and column $j$ is denoted by $h(I)_{i j}$. The masked threshold, denoted as $t$, is set to $0.5$ in this article. Notably, the cost associated with generating auxiliary data is acceptable as it involves a one-time consumption cost.
\subsubsection{Saliency-aware Knowledge Transfer}
In order to leverage the rich prior knowledge, knowledge distillation is a natural way. Moreover, according to the Deutsch–Norman late-selection filter model, critical information selection occurs exclusively after a comprehensive analysis of all inputs at a higher level. Inspired by this, this article provides high-level supervision to enable the model to concentrate on objects during prediction, effectively mitigating the adverse effects of background clutter. As illustrated in Fig.~\ref{fig:sgml-net}, this paper introduces a  symmetric structure of the main branch. Considering the saliency as the input for the additional branch, the resulting feature representation mapped to the latent space distinctly assumes an object-focused nature. While this feature exhibits an object-focused characteristic, transferring guidance from the feature dimension poses challenges, particularly when data availability is limited. Hence, this paper proposes to offer guidance from the standpoint of prediction distribution. The model prioritizes object relationships by minimizing the discrepancy between the distribution generated from raw images $p_{I}$ and that produced from saliency $p_{S}$ . The discrepancy between probability distributions can be measured using KL divergence as follows:
\begin{equation}
    D_{KL}(p_{I}||p_{S}) = \sum_{i} p_{I}(i) \log\frac{p_{I}(i)}{p_{S}(i)}.
\end{equation}

Nevertheless, employing traditional unidirectional supervision may not be optimal due to the large gap in representation between the saliency and the raw data and the potential risk of error amplification. After carefully considering these factors, this article utilized the deep mutual learning paradigm~\cite{zhang2018deep} to align the probabilities between the image and foreground bidirectionally. The loss function for saliency-aware guidance can be defined as follows:
\begin{equation}
    \mathcal{L}_{\text {SaG }}= D_{K L}\left(p_{I} \| p_{S}\right) +  D_{K L}\left(p_{S} \| p_{I}\right).
    \label{eq:mutual}
\end{equation}

Notably, the article applies the same data augmentation to both images and their corresponding prior before feeding them into the network to ensure that the input difference remains within an acceptable range and maintains consistency in the probability.

\subsection{Representation Highlight\&Summarize (RHS) Module}
The saliency-aware guidance described in the previous section facilitates the network to learn object-focused information within the distribution perspective. However, it provides limited guidance for attention towards the representation produced by the backbone network. Moreover, the quality of representation also significantly impacts the effectiveness of guidance. To provide robust, deep information for fine-grained prediction, the RHS module is devised to simultaneously highlight and summarize crucial patterns within the representation level across multiple objects.
\subsubsection{Representation Highlight} 
Due to significant differences in posture and scale, the global representation obtained by common FSL may not be sufficient for FS-FGVR. Inspired by previous work~\cite{wu2021object,zha2023boosting}, different sets’ relationships are established to obtain informative representation. As illustrated in Fig.~\ref{fig:sgml-net}, the encoder $f_\theta(\cdot)$ produces high-level semantic representations for all images in the  $N$-way $K$-shot episode $\mathcal{T}$=$(\mathcal{D}^\mathcal{S}, \mathcal{D}^\mathcal{Q})$. Next, the semantic representations of each class in $\mathcal{D}^\mathcal{S}$ are aggregated to compute the prototypes by the following formula:
\begin{equation}
     {pt}_i=\frac{1}{\left| \mathcal{D}_{\text {i}}^{\mathcal{S}}\right|} \sum_{x \in  \mathcal{D}_{\text {i}}^{\mathcal{S}}} f_\theta(x),
\end{equation}
where $p t_i$ represents the prototype of the \textit{i-th} class and $\mathcal{D}_{\text {i}}^{\mathcal{S}}$ refers to the \textit{i-th} class within the support set. Then, the RHS module applies $1\times1$ convolution to project the prototype feature map $pt$ and query feature map  $f^q$ to key $k$ and query $q$, respectively, to filter out redundant information. Then, the semantic relation between $k$ and $q$ is explored. The relation matrix $M$ between $q$ and $k$ can be calculated using the following formula:
\begin{equation}
    M_{i j}(q,k)=\operatorname{Softmax}\left(\left\langle q_i^T, k_j\right\rangle\right),
    \label{eq:relation}
\end{equation}
where $q_i \in \mathbb{R}^{c \times 1}$ and $k_j\in \mathbb{R}^{c \times 1}$ are the local descriptor of $q$ and $k$, respectively, and 
$M \in \mathbb{R}^{hw \times hw}$. $<,>$ denotes the metric function.  This article chooses cosine similarity as the metric function due to its superior ability to measure local representation similarity.

The prototype feature representation can be highlighted by computing a linear combination of the first-order and second-order representations by using the following equation:
\begin{equation}
     \overline{pt}= \hat{k} \cdot M+\hat{pt},
     \label{eq:combine}
\end{equation}
where $\overline{pt}$ denotes the refined prototype feature representation. The first-order representation is represented by $\hat{pt}^s  \in \mathbb{R}^{c \times hw}$, which is the spatially vectorized form of $pt$. The second-order representation is obtained by rearranging the first-order representation using a relation matrix $M$. 

\subsubsection{Representation Summarize}
 Existing research is mainly based on high-dimensional local descriptors, which may present a potential overfitting concern. Considering the potential risk for overfitting and computational efficiency, the RHS summarizes the highlighted information into contextual embedding. The RHS employs a combination of global max pooling and global average pooling to aggregate features while retaining the dominant feature, which can be expressed as
\begin{equation}
    \left\{\begin{array}{l}
\tilde{pt}=\frac{1}{h \times w} \sum_{i \in h w} \bar{pt}_i+\max _{i \in h w} \tilde{pt}_i, \\
\tilde{f}^q=\frac{1}{h \times w} \sum_{i \in h w} f_i^q+\max _{i \in h w} f_i^{q},
\end{array}\right.
\label{eq:pooling}
\end{equation}
where $\widetilde{pt}$ and $\widetilde{f}^q$ are the final refined representation of prototype feature representation $pt$ and query feature representation $f^q$.

\subsection{Overall Learning Objective}
Recalling the objectives for an FS-FGVR algorithm, the optimization problem arises as 
\begin{equation}
    \mathcal{L}_{\text {total }}=\mathcal{L}_{cls1}+\mathcal{L}_{cls2}+\alpha \mathcal{L}_{\text {SaG }},
\end{equation}
where $\mathcal{L}_{cls1}$ and $\mathcal{L}_{cls2}$ are the classification losses for two branches, respectively, calculated as follow:

\begin{equation}
    P\left(y=c \mid x ; \mathcal{D}^\mathcal{S}\right)=\frac{\exp \left(\left\langle\tilde{f}^q, \tilde{pt} _c\right\rangle\right)}{\sum_{n \in N} \exp \left(\left\langle\tilde{f}^q, \tilde{p t}_n\right\rangle\right)},
\end{equation}

\begin{equation}
    \mathcal{L}_{c l s}=\sum_{(x, y) \in \mathcal{D}^\mathcal{Q}}-y \log P\left(y=c \mid x ; \mathcal{D}^\mathcal{S}\right),
\end{equation}
where  $<,>$ is the metric function. The article selects cosine similarity as the metric function in the probability computation to ensure consistency with the metric space of the RHS module. $\alpha$ is the hyper-parameter, which controls the interaction degree of saliency-aware guidance. 

\section{PERFORMANCE EVALUATION}
\label{exp_result}
The benchmarks are introduced in Sec.~\ref{datasets}, followed by the detailed implementation of the proposed approach in Sec.~\ref{implementation}. The results of comparison experiments with different methods are presented in Sec.~\ref{compare_sota},~\ref{compare_sg_ml}, and~\ref{compare_backbone}, respectively.

\subsection{Datasets}
\label{datasets}
This article evaluates the performance of the proposed approach on three commonly used fine-grained benchmarks, including CUB-200-2011\cite{wah2011caltech}, Stanford Dogs\cite{khosla2011novel} and  Stanford Cars\cite{krause20133d}. Details of those benchmarks are described below:
\begin{itemize}
		\item {\bf CUB-200-2011} consists of $11,788$ images depicting $200$ species of birds. Following the setup of~\cite{zha2023boosting}, the article randomly assigns $100$, $50$, and $50$ classes to the training, validation, and testing. To ensure a fair comparison, none of the methods compared in this study utilize the bounding-box annotations provided by this benchmark.
		\item {\bf Stanford Dogs} contains $20,580$ images depicting  $120$ species of dogs. Following the setup of~\cite{zha2023boosting}, this article randomly ad assigns $70, 20$, and $30$ classes for training, validation, and testing, respectively.
		\item {\bf Stanford Cars} comprises $16,185$ images depicting $196$ car models. Following the setup of~\cite{zha2023boosting}, this article randomly assigns $130, 17$, and $49$  classes to the training, validation, and testing, respectively.
\end{itemize}

\subsection{Implementation Details}
\label{implementation}

\textbf{Network architecture:} Following the previous research \cite{zha2023boosting}, this article employs ResNet-12 as the primary backbone to evaluate model performance. ResNet-12 consists of four residual blocks, each with three convolutional layers utilizing a $3\times3$ kernel and a  $2\times2$ max-pooling layer. Unlike previous studies~\cite{zhou2021binocular,xu2021variational} that employed wider-width ResNet-$12$ with a drop-block as a regularizer, this study uses a ResNet-$12$ with filter numbers $64$-$128$-$256$-$512$ without introducing the drop-block to ensure a fair comparison. Moreover, the model's performance on the commonly-used shallower backbone Conv4 in FS-FGVR is discussed in Section \ref{compare_backbone}.

\textbf{Pre-training Stage:} 
Some recent work~\cite{chen2019closerfewshot,wang2019simpleshot,chen2021meta} observed that pre-trained models for whole-classification have improved the transferability of novel classes in meta-learning models. Currently, almost all state-of-the-art methods\cite{xu2021variational,wu2021object,zhang2022learning} utilize a two-stage approach involving pre-training and episodic training for training in the context of FS-FGVR tasks. This article follows the pre-training setting in~\cite{ye2020few,chen2021meta}. During the pre-training stage, a fully-connected layer is appended at the end of the backbone for whole-class classification. All the images apply standard data augmentation, including random crop and random flip. 
For ResNet-$12$, the study utilizes the SGD optimizer with Nesterov acceleration, incorporating a learning rate of $0.001$, weight decay of $0.0005$, and momentum of $0.9$. The model is trained from scratch for $300$ epochs, with the learning rate decreasing by $0.1$ at epoch $75$ and epoch $150$. As for Conv$4$, this study employs the SGD optimizer with a learning rate of $0.001$, momentum of $0.9$, and weight decay of $0.0005$. The model is trained from scratch for $200$ epochs, with the learning rate decreasing by $0.1$ at epoch $85$ and epoch $170$.

\textbf{Episodic Training Stage:}
After pre-training, the pre-trained model removed the last fully connected layer and fine-tuned the model following the standard meta-learning scheme. For CUB-200-2011 and Stanford Dogs under ResNet-$12$ backbone, this article utilizes the Adam optimizer with an initial learning rate of $0.001$ during the episodic training stage and optimizes the model over 40,000 episodes, except for the $5$-way $1$-shot setting, which involves $60000$ episodes. For the rest of the cases, this article employed the AdamW optimizer with an initial learning rate of $0.001$ during the episodic training stage. This article optimized the proposed model over $40000$ episodes in a $15$-way $5$-shot setting, evaluating its performance in both $1$-shot and $5$-shot settings for training stability.

\subsection{Comparison With State-of-the-Art Approaches}
\label{compare_sota}
The following experiments present a comparison of the proposed network with state-of-the-art approaches, comprising four generic FSL methods (ProtoNet\cite{snell2017prototypical}, DN4\cite{li2019revisiting},  Baseline++~\cite{chen2019closerfewshot}, CAN\cite{hou2019cross}) and five specialized FS-FGVR methods (TOAN~\cite{huang2021toan}, BSNet~\cite{li2020bsnet}, OLSA~\cite{wu2021object}, AGPF~\cite{TANG2022108792} and BSFA~\cite{zha2023boosting}). The performance evaluation is conducted on three fine-grained benchmarks mentioned in Sec.~\ref{datasets}. The performance evaluations of the proposed framework on different benchmarks are presented in Sec.~\ref{compare_in_cub}, Sec.~\ref{compare_in_dogs}, and Sec.~\ref{compare_in_cars}, respectively. 

\begin{table*}[!htbp]
	\centering
	\caption{This table reports the accuracy (\%) of $5$-way $1$-shot and $5$-shot tasks on three popular benchmarks using ResNet-12. The accuracy refers to the mean value with 95\% confidence intervals on $600$ episodes. The best results are presented in bold. ${*}$ indicates that those results are obtained from the original paper, while ${\dagger}$ denotes that the results are reported in \cite{zha2023boosting}. The remaining results are reproduced using the open-source code under the same experimental settings.}
	\resizebox{0.95\linewidth}{!}{ 
		\begin{tabular}{lccccccc}
			\addlinespace
			\toprule
			\specialrule{0em}{1pt}{1pt}
			\multirow{2}{*}{ \bf Method} &\multirow{2}{*}{ \bf Backbone} &\multicolumn{2}{c}{\bf CUB-200-2011} & \multicolumn{2}{c}{\bf Stanford Dogs} & \multicolumn{2}{c}{\bf Stanford Cars} \\ 
			\cline{3-8}
			\specialrule{0em}{1pt}{1pt}
			& &{$1$-shot} & {$5$-shot}& {$1$-shot} &{$5$-shot}& {$1$-shot} &{$5$-shot}\\
			\midrule
			ProtoNet~\cite{snell2017prototypical} &ResNet-12& {73.52} {\scriptsize $\pm$ 0.98} & {87.57} {\scriptsize $\pm$ 0.52} & {61.46} {\scriptsize $\pm$ 0.94}& {83.03} {\scriptsize $\pm$ 0.58} & {75.46} {\scriptsize $\pm$ 0.89}& {91.18} {\scriptsize $\pm$ 0.41} \\
			\specialrule{0em}{1pt}{1pt}

			DN4 $^{\dagger}$~\cite{li2019revisiting}  &ResNet-12& {64.95}  {\scriptsize $\pm$ 0.99} & {83.18} {\scriptsize $\pm$ 0.62} &  {49.70} {\scriptsize $\pm$ 0.85}  & {71.59} {\scriptsize $\pm$ 0.68}  & {75.79} {\scriptsize $\pm$ 0.84} & {94.14} {\scriptsize $\pm$ 0.35} \\
			\specialrule{0em}{1pt}{1pt}
		
			Baseline++~\cite{chen2019closerfewshot}  &ResNet-12& {72.07}  {\scriptsize $\pm$ 0.91} & {86.16} {\scriptsize $\pm$ 0.54} &  {64.63} {\scriptsize $\pm$ 0.82}  & {82.37} {\scriptsize $\pm$ 0.56}  & {76.14} {\scriptsize $\pm$ 0.82} & {91.15} {\scriptsize $\pm$ 0.40} \\
			\specialrule{0em}{1pt}{1pt}
        	CAN $^{\dagger}$~\cite{hou2019cross} &ResNet-12&{76.98}  {\scriptsize $\pm$ 0.48} & {87.77} {\scriptsize $\pm$ 0.30} &  {64.73} {\scriptsize $\pm$ 0.52}  & {77.93} {\scriptsize $\pm$ 0.35}  & {86.90} {\scriptsize $\pm$ 0.42} & {93.93} {\scriptsize $\pm$ 0.22}  \\

         	\specialrule{0em}{1pt}{1pt}
			TOAN$^{*}$~\cite{huang2021toan}   &ResNet-12& {66.10} {\scriptsize $\pm$ 0.86}  & {82.27} {\scriptsize $\pm$ 0.60}  & {49.77} {\scriptsize $\pm$ 0.86}  & {69.29} {\scriptsize $\pm$ 0.70}  & {75.28} {\scriptsize $\pm$ 0.72} & {87.45} {\scriptsize $\pm$ 0.48}\\
			\specialrule{0em}{1pt}{1pt}
			BSNet$^{\dagger}$~\cite{li2020bsnet}   &ResNet-12& {73.48} {\scriptsize $\pm$ 0.92}  & {83.84} {\scriptsize $\pm$ 0.59} & {61.95} {\scriptsize $\pm$ 0.97} & {79.62} {\scriptsize $\pm$ 0.63} & {71.07} {\scriptsize $\pm$ 1.03} & {88.38} {\scriptsize $\pm$ 0.62} \\
			\specialrule{0em}{1pt}{1pt}
			OLSA$^{*}$~\cite{wu2021object}   &ResNet-12& {77.77} {\scriptsize $\pm$ 0.44} & {89.87} {\scriptsize $\pm$ 0.24} & {64.15} {\scriptsize $\pm$ 0.49} & {78.28} {\scriptsize $\pm$ 0.32} & {77.03} {\scriptsize $\pm$ 0.46} & {88.85} {\scriptsize $\pm$ 0.46} \\
			\specialrule{0em}{1pt}{1pt}
			AGPF$^{\dagger}$~\cite{TANG2022108792}  &ResNet-12& {78.73} {\scriptsize $\pm$ 0.84} & {89.77} {\scriptsize $\pm$ 0.47}& {72.34 {\scriptsize $\pm$ 0.86}} &  {84.02 {\scriptsize $\pm$ 0.57}} & {85.34} {\scriptsize $\pm$ 0.74} & {94.79} {\scriptsize $\pm$ 0.35} \\
            \specialrule{0em}{1pt}{1pt}
			BSFA~\cite{zha2023boosting}   &ResNet-12&{82.22} {\scriptsize $\pm$ 0.85} & {90.49} {\scriptsize $\pm$ 0.47} & {69.62} {\scriptsize $\pm$ 0.92} & {82.50} {\scriptsize $\pm$ 0.58} & \bf{89.42} {\scriptsize $\pm$ 0.68} & \bf{95.36} {\scriptsize $\pm$ 0.36} \\
			\specialrule{0em}{1pt}{1pt}
             \midrule
			RSaD   &ResNet-12&   \bf {82.45 {\scriptsize $\pm$ 0.79}} & \bf {92.02 {\scriptsize  $\pm$ 0.44}} & \bf {73.75} {\scriptsize  $\pm$ 0.93} & \bf{86.65} {\scriptsize  $\pm$ 0.54} &  {{87.27} {\scriptsize$\pm$ 0.70}} & {95.01 {\scriptsize  $\pm$ 0.49}}  \\ \bottomrule
 	\end{tabular}}	
        \label{tb:comparison}
	\vspace{-2mm}
\end{table*}

\subsubsection{{Results on the CUB-200-2011 Dataset}}
\label{compare_in_cub}
As presented in Table~\ref{tb:comparison}, the RSaD outperforms FSL methods, and FS-FGVR methods on the CUB-$200$-$2011$ dataset. Specifically, RSaD achieves comparable performance with state-of-the-art method BSFA in $5$-way $1$-shot and $5$-way $5$-shot tasks, respectively. This indicates its ability to capture class correlations at various granularities effectively. One key factor contributing to the superior performance of RSaD lies in its ability to provide saliency guidance and region relations, resulting in better embedding spaces. In contrast to the proposed approach, BSFA incorporates a crop function aimed at discerning the foreground of images in both the training and testing phases. The crop function learned under the ResNet-$12$ backbone may be biased, potentially contributing to BSFA's inferior performance relative to the proposed approach.
\subsubsection{{Results on the Stanford Dogs Dataset}}
\label{compare_in_dogs}
In order to evaluate the capability of RSaD in handling more challenging FS-FGVR tasks, this section further presents its performance on the more complex dataset Stanford Dogs. As shown in Table~\ref{tb:comparison}, RSaD outperforms state-of-the-art FS-FGVR approaches by a substantial margin in the Stanford Dogs dataset. Specifically, RSaD achieves $4.54\%$, $1.03\%$, $9.22\%$, $11.42\%$ and $23.6\%$ performance gain for the $1$-shot setting and $3.45\%$, $2.06\%$, $7.80\%$, $6.46\%$ and $16.79\%$ performance gain for the $5$-shot setting compared with the BSFA and four other FS-FGVR methods, respectively. An explanation for the remarkable performance of RSaD could be its ability to mitigate noise in complex environments, such as the background. This enables the model to allocate increased attention toward the intrinsic dissimilarities within the image.
\subsubsection{{Results on the Stanford Cars Dataset}}
\label{compare_in_cars}
To further evaluate the efficacy of the RSaD, this article conducted experiments on the simpler benchmark, Stanford Cars. As depicted in Table~\ref{tb:comparison}, RSaD achieves competitive results compared to state-of-the-art methods. Specifically, Our method outperforms all typical FSL methods and some FS-FGVR methods under 1-shot and 5-shot. However, RSaD performs slightly worse than the state-of-the-art methods BSFA by 2.15\% for $1$-shot, and 0.35\% for $5$-shot, respectively. Their superior performance may be attributed to their additional structure and spatial descriptor comparison. Considering the trade-off between computational efficiency and performance, RSaD avoids incorporating excessive additional structure. Furthermore, RSaD employs spatial descriptor aggregation to avoid cost-expensive spatial descriptor comparison. Undoubtedly, RSaD provides significant computational efficiency advantages, as shown in Sec.~\ref{tb:complexity} for details. 

\subsection{Comparison With Saliency-Guided Approaches and KD-Based Approaches}
As the proposed approach is a hybrid of saliency-guided and knowledge-distillation (KD) based approaches, this sub-section compares RSaD with the saliency-guided and KD-based methods. The comparison results are presented in Sec.~\ref{sec:SGA} and Sec.~\ref{sec:KD}, respectively.
\label{compare_sg_ml}

\begin{table}[t]
\caption{Comparison with other few-shot knowledge distillation based methods and saliency-guided approach on CUB-200-2011.The best results are shown in bold. ${*}$ indicates that those results are obtained from the original paper.}
\centering
\begin{tabular}{lccc}
\toprule
\multirow{2}{*}{Method} &\multirow{2}{*}{Backbone}& \multicolumn{2}{c}{CUB-200-2011}\\
\cmidrule(lr){3-4}
~&~&5-way 1-shot&5-way 5-shot\\
\midrule 
FOT$^{*}$\cite{wang2021fine}&ResNet-18&{80.40}&{89.68}\\
SGCA$^{*}$\cite{zhao2021saliency}&ResNet-12& {79.84 {\scriptsize $\pm$ 0.42}}& {90.91 {\scriptsize $\pm$ 0.22}}\\
SGCA++$^{*}$\cite{zhao2022boosting}&ResNet-12& {80.69 {\scriptsize $\pm$ 0.42}}& {91.43 {\scriptsize $\pm$ 0.22}}\\
\midrule 
Protonet\cite{snell2017prototypical}&ResNet-12&{73.52 {\scriptsize $\pm$ 0.98}}&{87.57  {\scriptsize $\pm$ 0.52}}\\
Meta-Basline\cite{chen2021meta}&ResNet-12& {76.49 {\scriptsize $\pm$ 0.87}}& {87.86 {\scriptsize $\pm$ 0.52}} \\
BML$^{*}$\cite{zhou2021binocular}&ResNet-12& {76.21 {\scriptsize $\pm$ 0.63}}&  {90.45 {\scriptsize $\pm$ 0.36}}\\
LST-Proto$^{*}$\cite{ye2022few}&ResNet-12& {75.80 {\scriptsize $\pm$ 0.21}}& {90.22 {\scriptsize $\pm$ 0.12}}\\
LST-FEAT$^{*}$\cite{ye2022few}&ResNet-12& {80.20 {\scriptsize $\pm$ 0.21}}& {91.49 {\scriptsize $\pm$ 0.12}}\\
\midrule 
RSaD&{ResNet-12}&\textbf{82.45{ \scriptsize $\pm$ 0.79}}&\textbf{92.02 {\scriptsize $\pm$ 0.44}}\\

\bottomrule
\end{tabular}
\label{tb:SG_KD}
\end{table}
\subsubsection{{Comparison With Saliency-Guided Approaches}}
\label{sec:SGA}
This article compares RSaD with three other saliency-guided methods: SGCA~\cite{zhao2021saliency}, SGCA++~\cite{zhao2022boosting} and FOT~\cite{wang2021fine}. SGCA and SGCA++ obtain transferable representations through learning saliency-guided attention, while FOT synthesizes additional samples using a saliency map matching strategy. Table ~\ref{tb:SG_KD} presents the results. Compared to SGCA, RSaD achieves $2.61\%$ and $1.11\%$ performance gain in $1$-shot and $5$-shot tasks under backbone ResNet-$12$. Furthermore, RSaD outperforms SGCA, which is a refined version of SGCA. Compared to FOT, RSaD achieves $3.43\%$ and $2.45\%$ performance gain in $1$-shot and $5$-shot tasks under backbone ResNet-$18$. The superior performance of RSaD shows that the salient features obtained by RSaD are beneficial for learning a better embedding space.

\subsubsection{{Comparison With KD-Based FSL Approaches}}
\label{sec:KD}
This article compared RSaD with several recently published KD-based FSL approaches, including Meta-Baseline\cite{chen2021meta}, BML\cite{zhou2021binocular} and LST\cite{ye2020few}. Meta-Baseline employs a simple process of meta-learning over a whole-classification pre-trained model. BML utilizes mutual learning to align the feature distributions between local and global views. LST proposes a general learning strategy that allows the model to learn from the pre-trained strong classifier. Both their method and the proposed method are based on the typical FSL approach ProtoNet\cite{snell2017prototypical}, resulting in improved learning of the embedding space. LST-FEAT is an improved version of LST-Proto that applies LST with self-attention to the ProtoNet. Performance comparison is shown in Table~\ref{tb:SG_KD}. As can be seen, knowledge distillation can significantly improve the performance of FSL methods. The experiment results indicate that saliency prior may offer higher quality dark knowledge for knowledge distillation than using pre-trained models, potentially contributing to our model's superiority over other approaches. Moreover, even if pre-trained dark knowledge is optimized, unidirectional knowledge distillation may not be optimal in the $1$-shot scenario.

\subsection{Few-shot Fine-Grained Classification with Shallow Backbone Network}
\label{compare_backbone}
\begin{table*}[!htbp]
	\centering
	\caption{\label{tab:1} This table reports the accuracy (\%) of $5$-way $1$-shot and $5$-shot tasks on three popular benchmarks using Conv4. The best results are presented in bold. ${*}$ indicates that those results are obtained from the original paper. The remaining results are reproduced using the open-source code under the same experimental settings.}
	\resizebox{0.95\linewidth}{!}{ 
		\begin{tabular}{lccccccc}
			\addlinespace
			\toprule
			\specialrule{0em}{1pt}{1pt}
			\multirow{2}{*}{ \bf Method} &\multirow{2}{*}{ \bf Backbone} &\multicolumn{2}{c}{\bf CUB-200-2011} & \multicolumn{2}{c}{\bf Stanford Dogs} & \multicolumn{2}{c}{\bf Stanford Cars} \\ 
			\cline{3-8}
			\specialrule{0em}{1pt}{1pt}
			& &{$1$-shot} & {$5$-shot}& {$1$-shot} &{$5$-shot}& {$1$-shot} &{$5$-shot}\\
			\midrule
			\specialrule{0em}{1pt}{1pt}
             ProtoNet~\cite{snell2017prototypical}&Conv4& {51.78} {\scriptsize $\pm$ 0.93} & {74.72} {\scriptsize $\pm$ 0.69} & {41.60} {\scriptsize $\pm$ 0.82}& {56.98} {\scriptsize $\pm$ 0.75} & {46.38} {\scriptsize $\pm$ 0.79}& {64.58} {\scriptsize $\pm$ 0.67} \\
			\specialrule{0em}{1pt}{1pt}

           LRPABN$^{*}$ ~\cite{huang2020low}&Conv4& {63.63} {\scriptsize $\pm$ 0.77} & {76.06} {\scriptsize $\pm$ 0.58} & {45.72} {\scriptsize $\pm$ 0.75}& {60.94} {\scriptsize $\pm$ 0.66} & {60.28} {\scriptsize $\pm$ 0.76}& {73.29} {\scriptsize $\pm$ 0.63} \\
        \specialrule{0em}{1pt}{1pt}
           MattML$^{*}$ ~\cite{zhu2020multi}&Conv4& {66.29} {\scriptsize $\pm$ 0.56} & {80.34} {\scriptsize $\pm$ 0.30} & {54.84} {\scriptsize $\pm$ 0.53}& {71.34} {\scriptsize $\pm$ 0.38} & {66.11} {\scriptsize $\pm$ 0.54}& {82.80} {\scriptsize $\pm$ 0.28} \\
        \specialrule{0em}{1pt}{1pt}
           SoSN$^{*}$ ~\cite{zhang2019power}&Conv4& {64.56} {\scriptsize $\pm$ 0.91} & {77.82} {\scriptsize $\pm$ 0.57} & {48.21} {\scriptsize $\pm$ 0.72}& {63.15} {\scriptsize $\pm$ 0.67} & {62.88} {\scriptsize $\pm$ 0.72}& {76.10} {\scriptsize $\pm$ 0.28} \\
        \specialrule{0em}{1pt}{1pt}
	TOAN$^{*}$~\cite{huang2021toan}&Conv4& {65.34} {\scriptsize $\pm$ 0.75}  & {80.43} {\scriptsize $\pm$ 0.60}  & {49.30} {\scriptsize $\pm$ 0.77}  & {67.16} {\scriptsize $\pm$ 0.49}  & {65.90} {\scriptsize $\pm$ 0.72} & {84.24} {\scriptsize $\pm$ 0.48}\\
			\specialrule{0em}{1pt}{1pt}
           	BSFA~\cite{zha2023boosting} &Conv4& {59.40}  {\scriptsize $\pm$ 0.97} & {74.42} {\scriptsize $\pm$ 0.62} &  {49.13} {\scriptsize $\pm$ 0.84}  & {63.27} {\scriptsize $\pm$ 0.73}  & {56.06} {\scriptsize $\pm$ 0.89} & {73.28} {\scriptsize $\pm$ 0.68} \\
			\specialrule{0em}{1pt}{1pt}
            MLSO$^{*}$~\cite{zhang2022multi}&Conv4& {68.21} {\scriptsize $\pm$ 0.78} & {82.18} {\scriptsize $\pm$ 0.47} & {55.62} {\scriptsize $\pm$ 0.58}& {71.98} {\scriptsize $\pm$ 0.71} & \bf{67.83} {\scriptsize $\pm$ 0.63}& \bf{84.83} {\scriptsize $\pm$ 0.48} \\
        \specialrule{0em}{1pt}{1pt}
        \midrule
   	   RSaD &Conv4& \bf {71.15 {\scriptsize $\pm$ 0.92}} & \bf {84.03 {\scriptsize $\pm$ 0.62}} & \bf {59.42} {\scriptsize $\pm$ 0.95} & \bf{75.30} {\scriptsize  $\pm$ 0.69} &  {{65.43} {\scriptsize$\pm$ 1.29}} & {81.75 {\scriptsize  $\pm$ 0.88}} 
        \\ \bottomrule
 	\end{tabular}}	
        \label{tb:comparison_conv4}
	\vspace{-2mm}
\end{table*}
This article conducts experiments on shallower backbone Conv$4$ to further evaluate the model performance. Conv$4$ is composed of four blocks, each containing a $3\times3$ convolution layer, a batch normalization layer, a ReLU activation layer, and a $2\times2$ max-pooling layer. The experiments provide a thorough comparison of the proposed network with state-of-the-art approaches, including three FSL methods (ProtoNet~\cite{snell2017prototypical}, SoSN~\cite{zhang2019power}, MLSO\cite{zhang2022multi}), as well as four specialized FS-FGVR methods (LRPABN\cite{huang2020low}, MattML\cite{zhu2020multi}, TOAN\cite{huang2020low}, BSFA\cite{zha2023boosting}) using a Conv4 backbone. The results in Table~\ref{tb:comparison_conv4} demonstrate that the proposed approach has a performance advantage over shallow backbones. Due to the low capacity of Conv4, BSFA fails to generate optimal foreground object coordinates for background suppression, resulting in significant performance degradation. However, our proposed approach maintains superior performance despite the low-capacity backbone due to effective probability distribution comparison in a lower-dimensional space determined by the number of ways.  Overall, RSaD consistently performs well on multiple backbones, thus providing strong evidence of its effectiveness.

\section{Analysis And Discussion}
\label{discussion}
This section presents a series of ablation studies and visualizations to investigate each module's significance. Sec.~\ref{ablation} introduces the module-wise ablation experiment, while Sec.~\ref{impact_sp},~\ref{impact_fr}, and~\ref{imparct_mil} demonstrate the impact of saliency-aware guidance and RHS module. Additionally, Sec.~\ref{complexity} provides an analysis of the model complexity.
\subsection{Module-Wise Ablation Study}
\label{ablation}
Table~\ref{tb:ab2} displays the results of module-wise ablation experiments performed on ResNet-$12$. The baseline is a model obtained through meta-learning on a whole-classification pre-trained model. RHS refers to the representation highlight\&summarize module, while SaG stands for saliency-aware guidance strategy. The results in Table~\ref{tb:ab2} demonstrate the effectiveness of the SaG and RHS module on the FS-FGVR task. SaG gains more significant improvement than the SaG module, indicating that prior saliency provides more valuable information for optimizing the model than second-order features. Combining SaG and RHS modules further improves the model's performance. The main reason is that the RHS module optimizes representation and results in higher-quality supervision. In this way, the model leverages such supervision more effectively, improving overall performance.
\begin{table}[!htb]
\caption{Ablation experiments of the proposed method on CUB-200-2011 and Stanford Dogs datasets under backbone ResNet-12. The best results are highlighted in bold.}
\centering
\setlength{\tabcolsep}{1.5mm}
\begin{tabular}{cccccc}
\toprule
 \multirow{2}{*}{SaG} & \multirow{2}{*}{RHS}& \multicolumn{2}{c}{CUB-200-2011} & \multicolumn{2}{c}{Stanford Dogs} \\
\cmidrule(lr){3-4} 	\cmidrule(lr){5-6}
&&1-shot &5-shot &1-shot&5-shot\\
\midrule 
  & &{77.58{\scriptsize $\pm$0.90}}&{88.24{\scriptsize $\pm$0.52}}&{69.40{\scriptsize$\pm$0.88}}&{84.42{\scriptsize $\pm$0.55}}\\
   \checkmark & &{80.28{\scriptsize $\pm$0.87}}&{91.18{\scriptsize $\pm$0.46}}&{72.03{\scriptsize $\pm$0.87}}&{85.42{\scriptsize $\pm$0.52}}\\
 &\checkmark&{78.50{\scriptsize $\pm$0.86}}&{88.87{\scriptsize $\pm$0.52}}&{72.66{\scriptsize $\pm$0.87}}&{86.13{\scriptsize $\pm$0.52}}\\
  \checkmark &\checkmark&\textbf {82.45{\scriptsize $\pm$0.79}}&\textbf {92.02{\scriptsize $\pm$0.44}}& \textbf{73.75{\scriptsize$\pm$0.93}}&\textbf {86.65{\scriptsize$\pm$0.54}}\\
\bottomrule
\end{tabular}   
\label{tb:ab2}
\end{table}

\subsection{Discussion on Impact of SaG strategy}
\label{impact_sp}
The SaG strategy plays a vital role in learning discriminative features in the proposed method. This strategy involves two primary stages: augmented saliency generation and saliency-aware knowledge transfer. This subsection conducts experiments and Grad-CAM~\cite{selvaraju2017grad} visualization on the CUB-200-2011 dataset to investigate the impact of in SaG strategy.

\subsubsection{Impact of Augmented Saliency Generation}
This article integrates two distinct saliency detection models (U2-Net~\cite{qin2020u2} and BASNet~\cite{qin2019basnet}) using an OR operation within augmented saliency generation to produce saliency maps. This subsection conducts quantitative comparisons to explore the individual contributions of these saliency detection models. As shown in Table~\ref{tb:different_source}, utilizing a single saliency detection model can yield high performance, illustrating that both models can effectively generate saliency from images. Furthermore, employing an ensemble strategy leads to consistently superior classification performance in all scenarios. This suggests that the ensemble approach offers a more dependable prior for model training, thus alleviating the performance constraints encountered by individual models in certain instances. 

\begin{table}[!htbp]
\caption{Quantitative evaluation of the augmented saliency generation strategy on CUB-200-2011 and Stanford Dogs dataset. The best results are in bold.}
\centering
\setlength{\tabcolsep}{0.5mm}
\begin{tabular}{ccccc}
\toprule
\multirow{2}{*}{Prior} & \multicolumn{2}{c}{CUB-200-2011} & \multicolumn{2}{c}{Stanford Dogs} \\
\cmidrule(lr){2-3} 	\cmidrule(lr){4-5}
&5-way 1-shot &5-way 5-shot &5-way 1-shot&5-way 5-shot\\
\midrule 
 U2-Net &{82.09{\scriptsize $\pm$0.81}}&{91.88{\scriptsize $\pm$0.42}}&{72.71{\scriptsize$\pm$0.88}}&{85.49{\scriptsize $\pm$0.53}}\\
 BASNet&{81.72{\scriptsize $\pm$0.79}}& {91.86{\scriptsize $\pm$0.43}}& {72.42{\scriptsize $\pm$0.87}}&{86.36{\scriptsize $\pm$0.50}}\\
 \midrule 
Ensemble&\bf{82.45{\scriptsize $\pm$0.79}}&\bf{92.02{\scriptsize $\pm$0.44}}& \bf{73.75{\scriptsize $\pm$0.93}}&\bf{86.65{\scriptsize $\pm$0.54}}\\
\bottomrule
\end{tabular}
\label{tb:different_source}
\end{table}

\begin{figure} [!htb]
		\centering
        \includegraphics[scale=0.46]{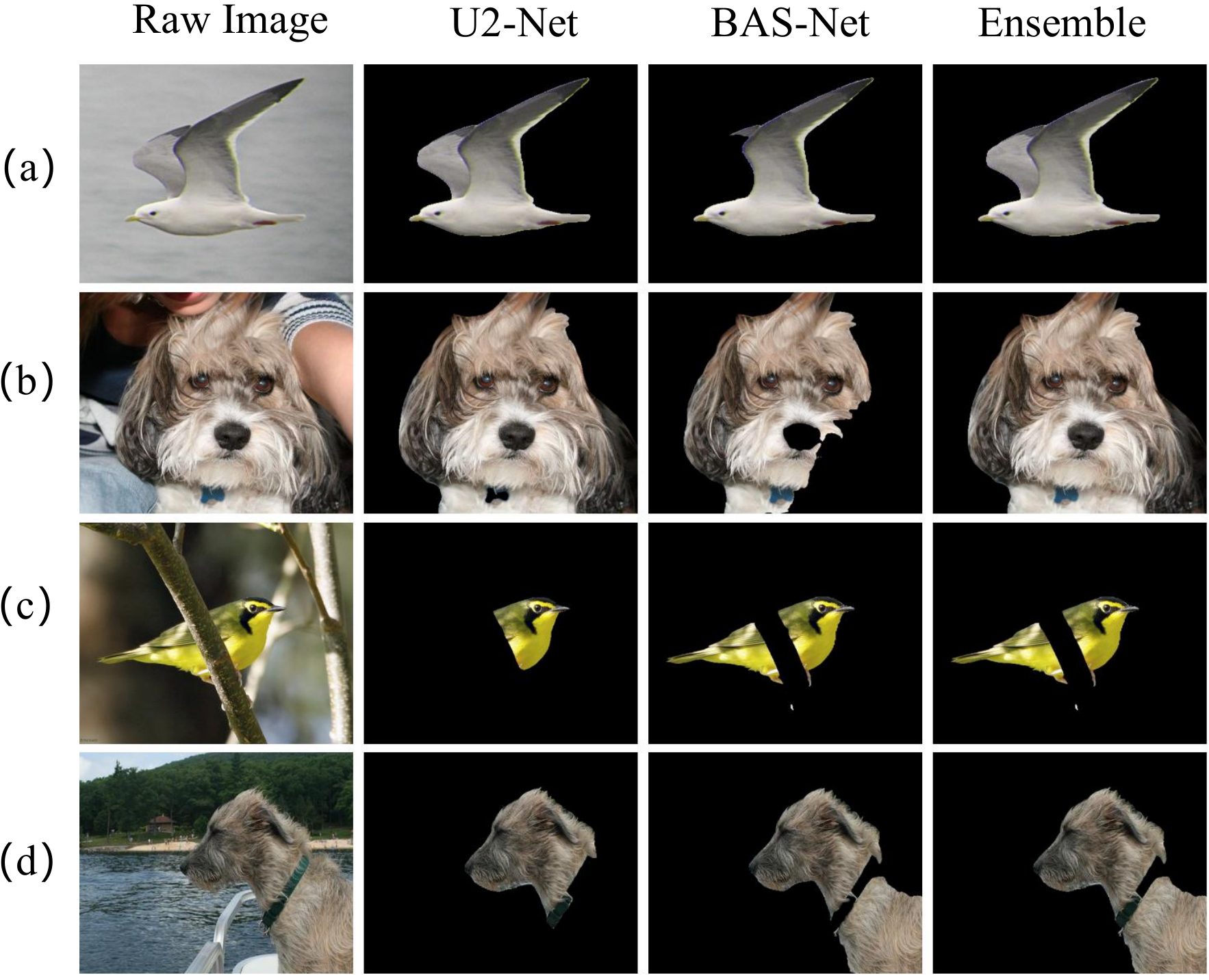}
		\caption{Visual comparison of the ensemble strategy with the individual model. The saliency produced by the ensemble model is more precise and more reliable than others.
}
		\label{fig:aog}
\end{figure}

Furthermore, this article presents a visualization of representative examples to demonstrate the efficacy of the ensemble strategy. As illustrated in Fig.~\ref{fig:aog}, BASNet may produce incomplete segmentation results for the whole object, while U2-Net typically performs robustly. This inconsistency may originate from  BASNet's reliance on a predict-refine architecture and the potential loss of local information due to dilated convolution\cite{Tang_2021_ICCV,ZHANG2022104423}. Conversely, BASNet tends to provide comprehensive segmentation in scenarios with occlusion, attributable to its utilization of hybrid boundary loss. Hence, the ensemble of the two models can produce more accurate saliency maps for further training.

\subsubsection{Impact of Saliency-aware Knowledge Transfer}
\begin{figure} [!htb]
		\centering
        \includegraphics[scale=0.4]{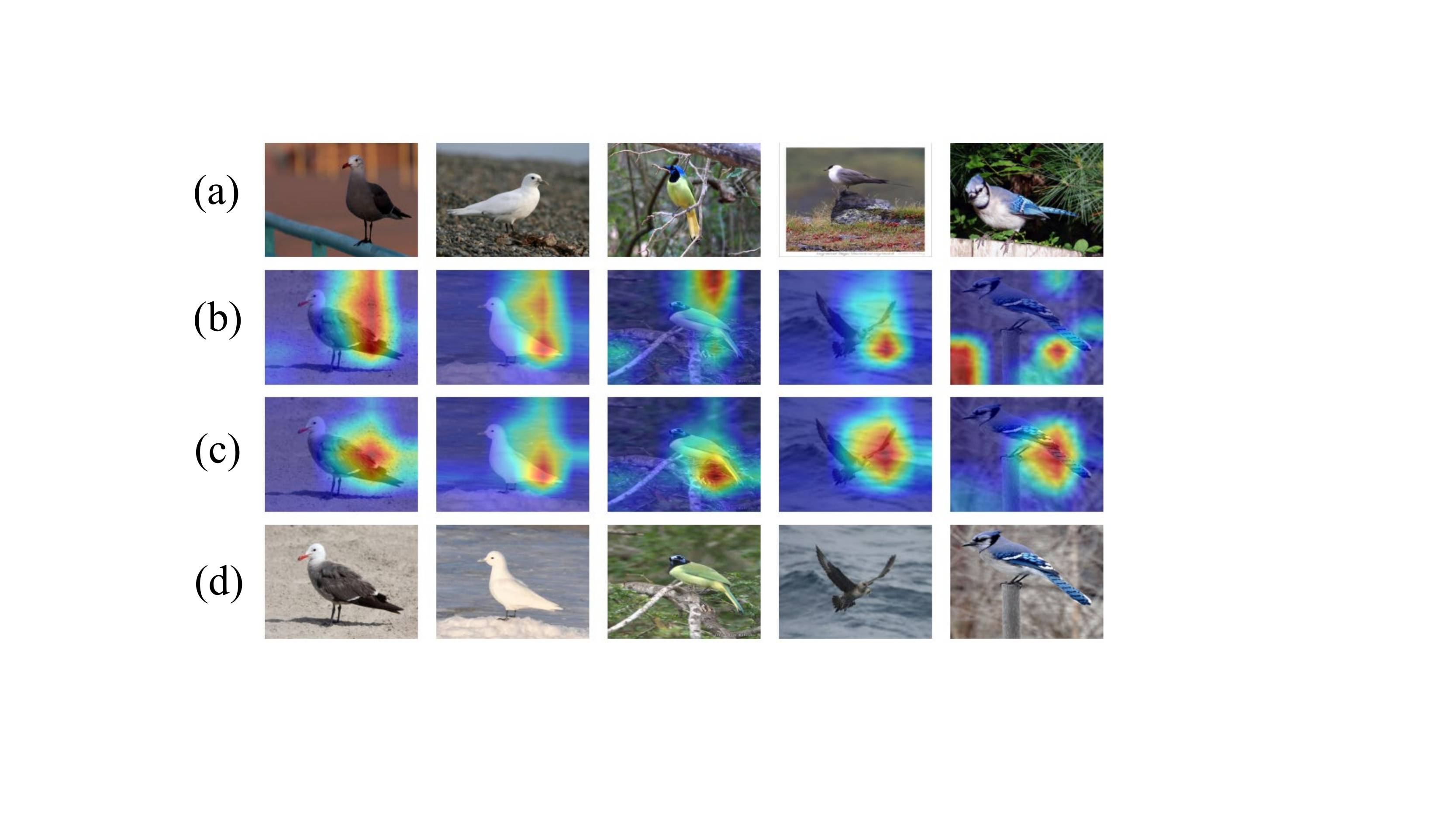}
		\caption{The visualization of local regions obtained from Grad-CAM. (a) prototype, (b) baseline, (c) our RSaD w/o RHS, and (d) query. The Grad-CAM Map identifies significant regions in the input image that affect classification decisions, where darker regions indicate higher importance. The specimens within each column belong to the same subclass, while those in different columns belong to distinct classes. Compared with baseline, RSaD pays more attention to the distinguishing characteristics of objects themselves.
}
		\label{fig:grad_cam}
\end{figure}

To further assess the impact of the SaG strategy, this article conducts Grad-CAM visualizations. Figs.5(b) and 5(c) visualize the class activation regions on the baseline and the proposed approach. The baseline model is obtained through meta-learning on a pre-trained classification model without using saliency as auxiliary information. The RHS module is excluded from the proposed framework in this study to eliminate any potential influence of the RHS module. Fig.~\ref{fig:grad_cam} illustrates that the class activation map generated by our proposed approach primarily focuses on the object itself while paying less attention to the background. In contrast, the baseline approach focuses on the part of the background in addition to the object itself. This indicates that introducing the SaG strategy can reduce interference from the background and capture the critical regions of the different sub-class. Thus, the robustness and generalization capabilities of the proposed model are improved. 

\subsection{Discussion on Impact of RHS module}
\label{impact_fr}

The RHS module is similar to the regular cross-attention module but lacks the value projection. To evaluate the effectiveness of the RHS module, this section conducts comparative experiments on the CUB-200-2011 and Stanford Dogs datasets. This study substituted the highlighted sub-module in the RHS with a cross-attention module. The results in Table\ref{tb:ca} indicate that the RHS approach outperforms the cross-attention mechanism, particularly in the 1-shot scenario. It is reasonable to assert that incorporating a value projection has the potential to introduce inconsistencies with the key projection, thereby leading to inaccurate alignment between prototypes and queries. Additionally, omitting a value projection enhances parameter efficiency, resulting in performance gain in the few-shot setup.

\begin{table} [!htb]
\caption{Comparison results of RHS module and cross attention. CA represents the cross-attention. The best results are presented in bold.}
\centering
\setlength{\tabcolsep}{1.0mm}
\begin{tabular}{lcccc}
\toprule
\multirow{2}{*}{Backbone}& \multicolumn{2}{c}{CUB-200-2011} & \multicolumn{2}{c}{Stanford Dogs} \\
\cmidrule(lr){2-3} 	\cmidrule(lr){4-5}
&1-shot &5-shot &1-shot&5-shot\\
\midrule 
RSaD w/o $RHS$&{80.28{\scriptsize $\pm$0.87}}&{91.18{\scriptsize $\pm$0.46}}&{72.03{\scriptsize$\pm$0.87}}&{85.42{\scriptsize $\pm$0.52}}\\
RSaD w/ $CA$&{81.58{\scriptsize $\pm$0.81}}& {91.53{\scriptsize $\pm$0.40}}& {72.47{\scriptsize $\pm$0.84}}&{86.50{\scriptsize $\pm$0.49}}\\
RSaD w/ $RHS$ &\bf{82.45{\scriptsize $\pm$0.79}}&\bf{92.02{\scriptsize $\pm$0.44}}& \bf{73.75{\scriptsize $\pm$0.93}}&\bf{86.65{\scriptsize $\pm$0.54}}\\

\bottomrule
\end{tabular}
\label{tb:ca}
\end{table}

The DBI (Davies-Bouldin Index) is employed as a quantitative metric to assess the quality of model embedding space. It is defined as follows:
\begin{equation}
    DBI = \frac{1}{n} \sum_{i=1}^{n} \max_{j \neq i} \Bigg( \frac{\sigma_i + \sigma_j}{d(c_i, c_j)} \Bigg),
\end{equation}
where $n$ represents the number of samples, $c_i$ denotes the cluster of the i-th sub-class, $\sigma_i$ signifies the average distance from the sample to the cluster in the i-th sub-class, and ${d(c_i, c_j)}$ represents the distance between different clusters. A smaller DBI value indicates a better clustering effect. As depicted in Table~\ref{tb:DBI}, applying RHS module can significantly improve the embedding space and the generalization performance in unseen subclasses. 

\begin{figure*}[!htb]
	\centering
	\begin{minipage}{0.22\linewidth}
		\centering
		\includegraphics[width=0.95\linewidth]{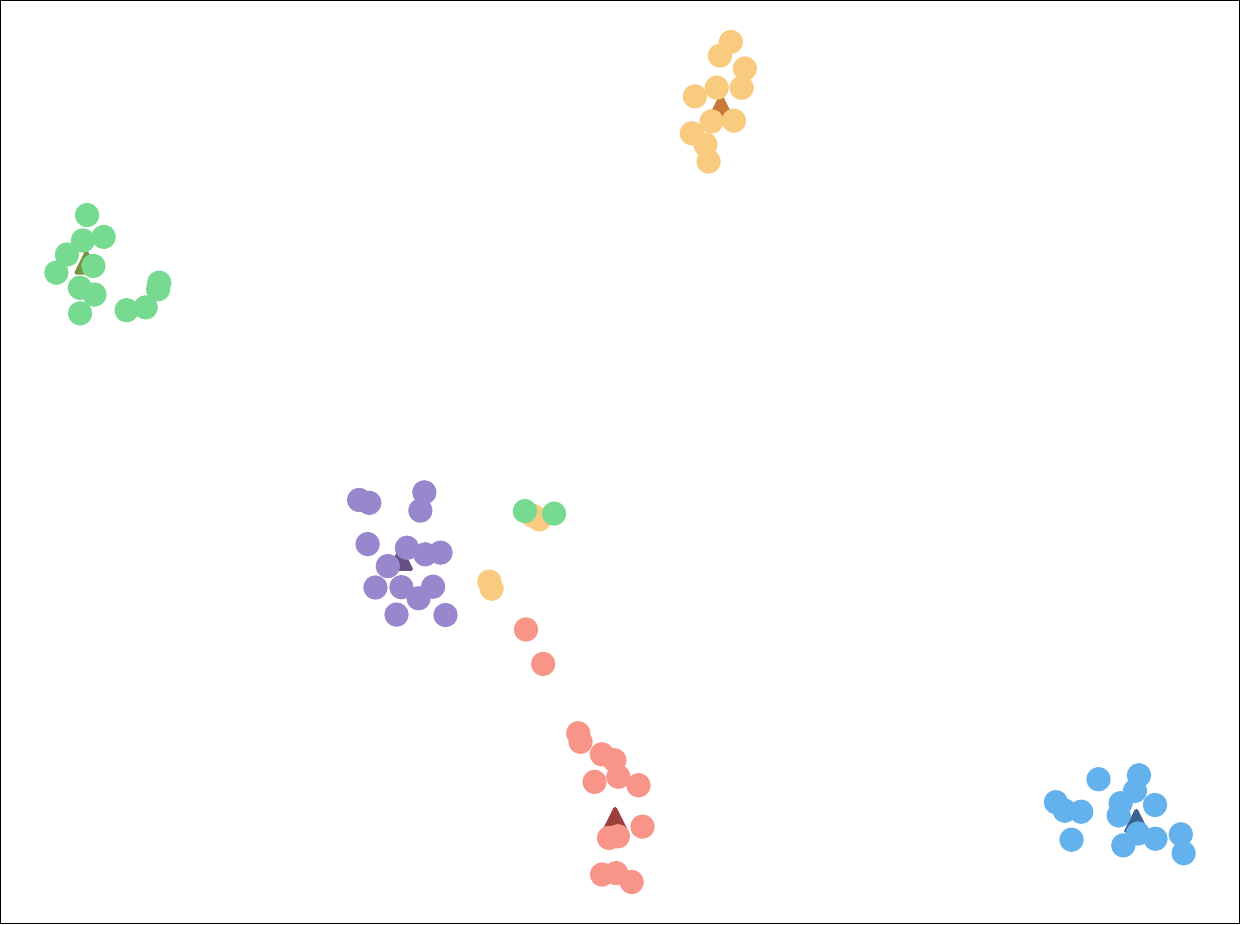}
		\centerline{(a)}
		\label{a}
		\vspace{0.1mm}
	\end{minipage}
	%\\qquad
	\begin{minipage}{0.22\linewidth}
		\centering
		\includegraphics[width=0.95\linewidth]{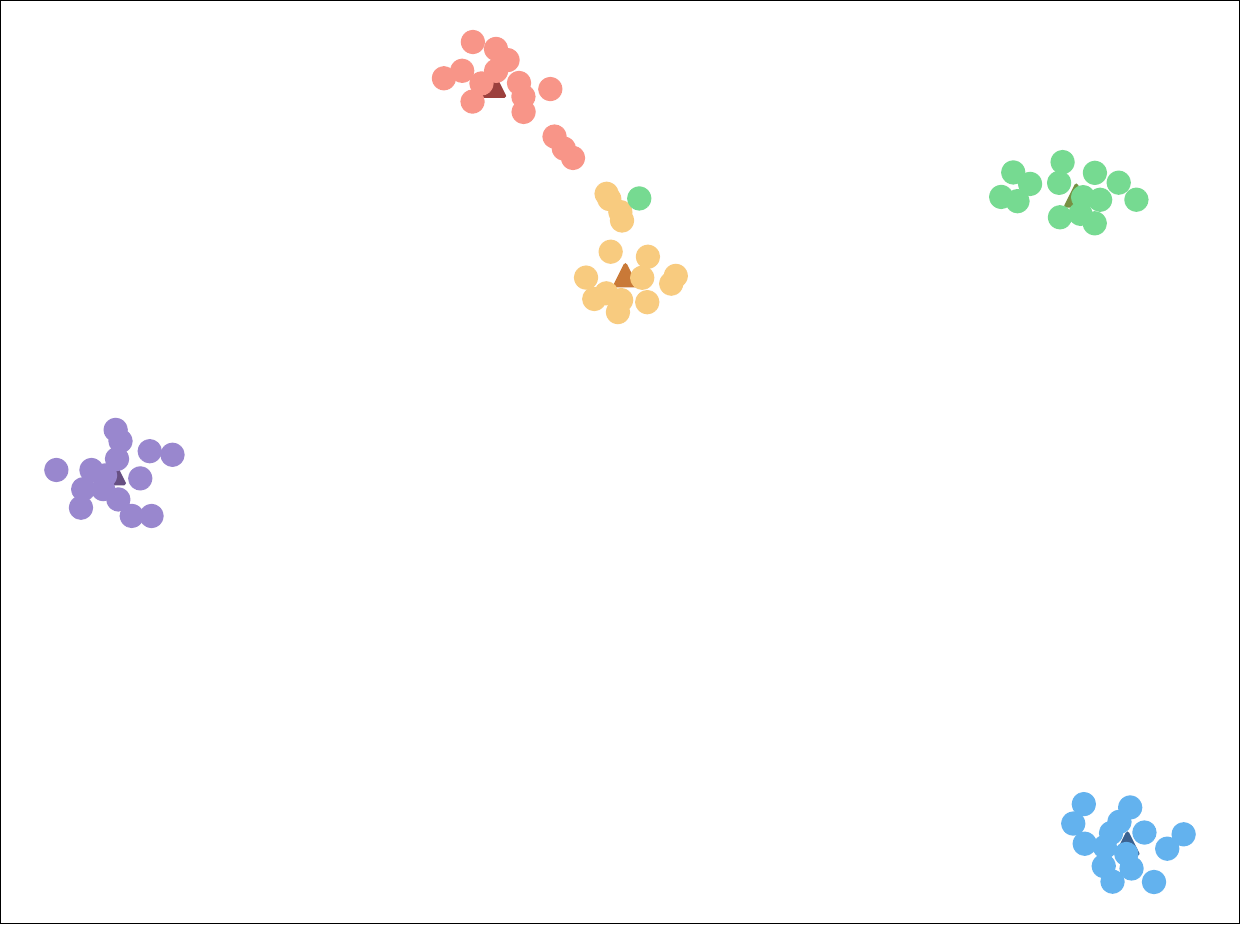}
		\centerline{(b) }
		\label{b}
		\vspace{0.1mm}
	\end{minipage}
	%\\qquad
	\begin{minipage}{0.22\linewidth}
		\centering
		\includegraphics[width=0.95\linewidth]{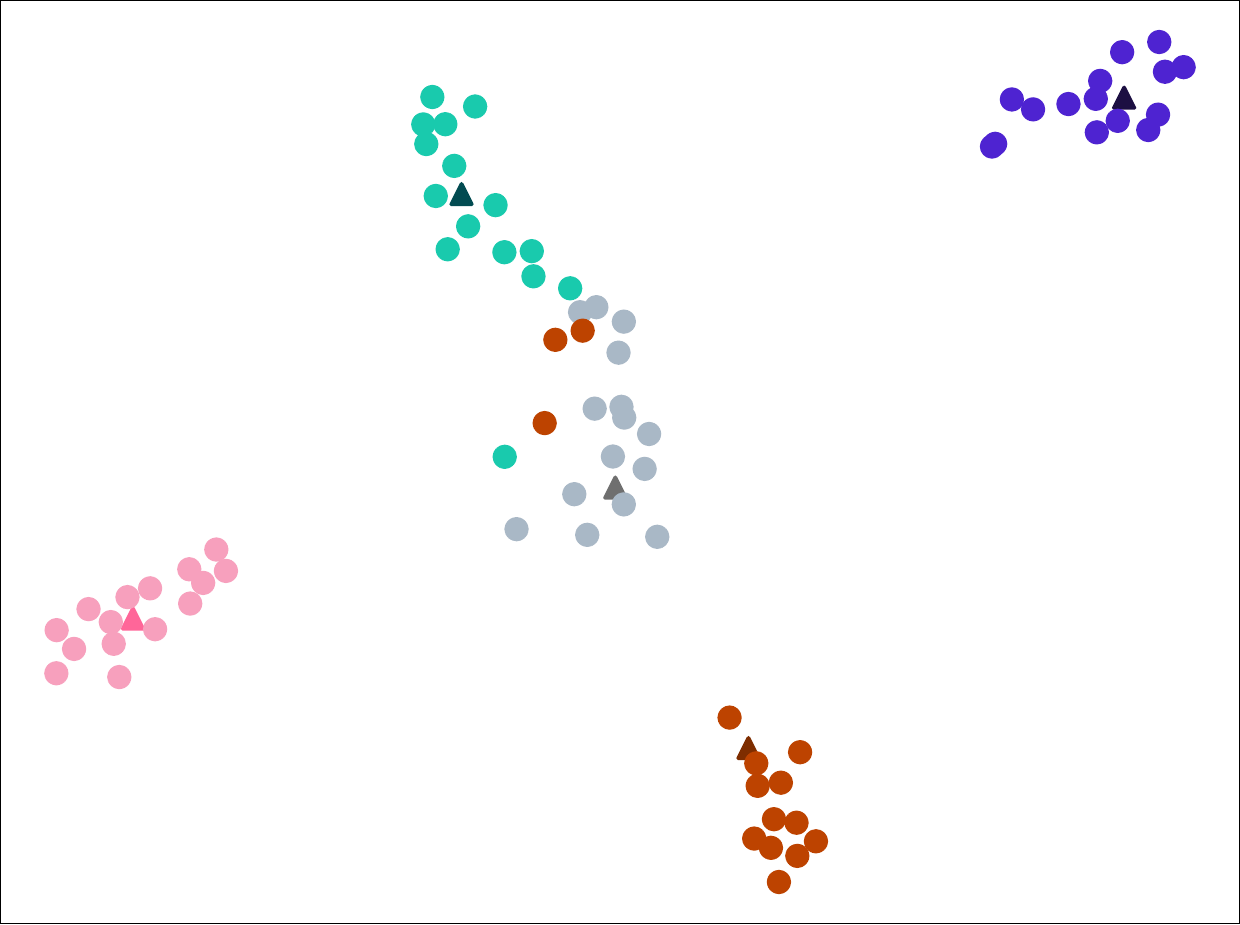}
		\centerline{(c) }
		\label{c}
		\vspace{0.1mm}
	\end{minipage}
	%\\qquad
	\begin{minipage}{0.22\linewidth}
		\centering
		\includegraphics[width=0.95\linewidth]{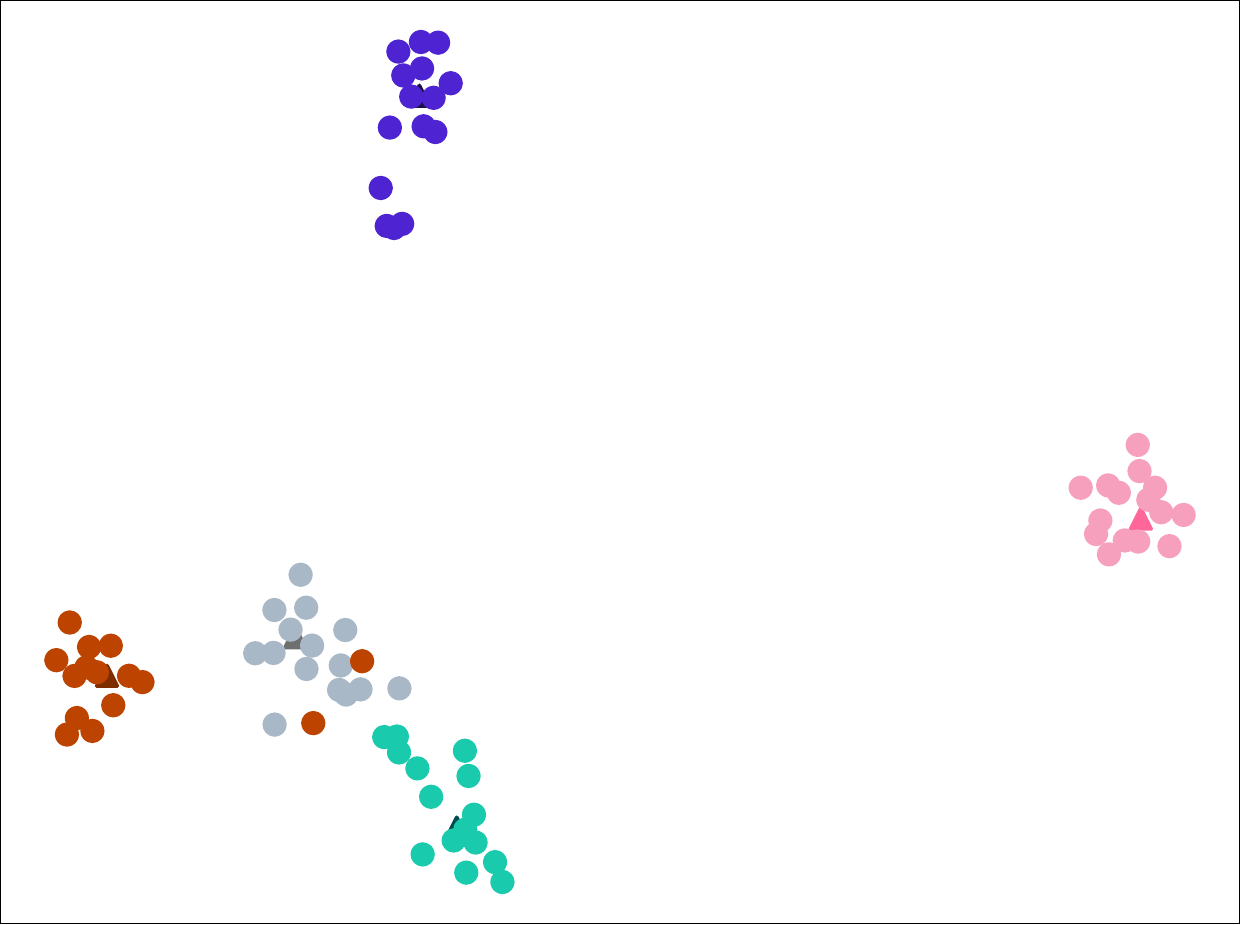}
		\centerline{(d)}
		\label{d}
		\vspace{0.1mm}
	\end{minipage}
	\caption{t-SNE visualization results for 5-way 1-shot tasks of different models on various sets in Dataset CUB-200-2011. (a) RSaD w/o RHS module on base set, (b) RSaD w/ RHS module on base set, (c) RSaD w/o RHS module on novel set, (d) RSaD w/ RHS module on novel set. Triangles represent prototypes, circles represent queries, and different colors indicate distinct subclasses. The labels of the base set and the novel set are disjoint.}
	\label{fig:tsne}
\end{figure*}

To further confirm the efficacy of the RHS module, t-SNE~\cite{van2008visualizing} visualization is performed. According to the results shown in Fig.~\ref{fig:tsne}, two observations are highlighted. Firstly, introducing the RHS module can significantly reduce the intra-class distance, resulting in a tighter cluster of objects within the same category. This leads to an improved discriminative capability of the model. Secondly, applying the RHS module can effectively enhance the model's generalization ability, resulting in consistently high performance in the novel set.

\begin{table}[!htb]
\caption{Quantitative experiment on the effect of RHS module. The terms "Base" and "Novel" indicate testing in the Base set and Novel set, respectively. The best results are presented in bold.}
\centering
\begin{tabular}{lccl}
\toprule
Method&Backbone&Split&DBI~$\downarrow$\\
\midrule 
RSaD w/o $RHS$ &ResNet-12&Base&3.7632\\
RSaD w/ $CA$ & ResNet-12&Base&3.0955 \\
RSaD w/ $RHS$ &ResNet-12&Base&\bf{2.9593} ($\downarrow 0.8102$)\\
\midrule 
RSaD w/o $RHS$ &ResNet-12&Novel&3.3140\\
RSaD w/ $CA$&ResNet-12&Novel&2.9642 \\
RSaD w/ $RHS$&ResNet-12&Novel&\bf{2.8374} ($\downarrow 0.4766$)\\
\bottomrule
\end{tabular}
\label{tb:DBI}
\end{table}

\subsection{Discussion on Impact of KD}
\label{imparct_mil}
 Considering the adoption of a symmetric structure and bidirectional distillation method in RSaD, it is reasonable to perform quantitative experiments to examine the effects of various distillation methods. Table~\ref{tb:ud_bd} presents the comparative results of various distillation methods, incorporating both directional and structural asymmetry. The "UD-KD" model is the unidirectional knowledge distillation model. The loss function is defined as follows:
 \begin{equation}
    \mathcal{L}_{\text {total }}=\mathcal{L}_{cls}+\alpha * D_{K L}\left(p_I \| p_S\right),
\end{equation}
where ${L}_{cls}$ is the loss function calculated by the cross-entropy, and $D_{K L}(,)$ represents the KL divergence. Here, $p_I$ denotes the logit distribution generated by the main branch, while $p_S$ represents the corresponding saliency logit distribution generated by the auxiliary branch. The auxiliary branch in "UD-KDP" is pre-trained, and its weights are frozen during training. The results demonstrate that RSaD outperforms the UD-KD model, suggesting that the bidirectional knowledge distillation strategy is better suited for distilling imperfect knowledge. Moreover, this study explores the performance of three different asymmetric branches on the CUB-200-2011 dataset: ResNet-12-Conv4 (strong-weak), ResNet-12-Visformer-tiny\cite{Chen_2021_ICCV} (similar parameters but different structure), and ResNet-12-ResNet-18 (weak-strong). Employing a symmetric branch structure demonstrates superior performance compared to using an asymmetric branch structure.

\begin{table}[!htb]
\caption{The effect of different mutual information distillation approaches. The best results are presented in bold.}
\centering
\begin{tabular}{llcc}
\toprule

\multirow{2}{*}{Method} & \multirow{2}{*}{Backbone} &\multicolumn{2}{c}{CUB-200-2011}\\
\cmidrule(lr){3-4}
~&~&5-way 1-shot&5-way 5-shot\\
\midrule 
UD-KD&ResNet-12/ResNet-12&{80.55{\scriptsize $\pm$0.81}}&{89.57{\scriptsize $\pm$0.48}}\\
UD-KDP&ResNet-12/ResNet-12&{81.02{\scriptsize $\pm$0.82}}&{91.00{\scriptsize $\pm$0.45}}\\
\midrule
RSaD&ResNet-12/Conv4&{81.70{\scriptsize $\pm$0.81}}&{91.60{\scriptsize $\pm$0.40}}\\
RSaD&ResNet-12/Visformer&{79.50{\scriptsize $\pm$0.84}}&{91.47{\scriptsize $\pm$0.42}}\\
RSaD&ResNet-12/ResNet-18&{79.72{\scriptsize $\pm$0.82}}&{90.88{\scriptsize $\pm$0.47}}\\
\midrule
RSaD&ResNet-12/ResNet-12&\bf{82.45{\scriptsize $\pm$0.79}}& \bf{92.02{\scriptsize $\pm$0.44}}\\
\bottomrule
\end{tabular}
\label{tb:ud_bd}
\end{table}
A parameter sensitivity analysis of the importance factor $\alpha$ is conducted to further explore the impact of saliency-aware supervision signal. According to the results shown in Table~\ref{tb:importance}, two insights are gained. Firstly, The degree of saliency-aware supervision signal is significant. Excessive saliency-aware supervision signal causes the model to overly focus on fitting the foreground distribution instead of the label distribution, while insufficient mutual information results in the model degrading to baseline performance. Secondly, the degree of saliency-aware supervision signalvaries across datasets. Considering the confidence of mutual information, less saliency-aware supervision signal may be more suitable for challenging datasets.
\begin{table} [!htb]
\caption{Effect of the degree of the saliency-aware supervision. The best results are presented in bold.}
\centering
\setlength{\tabcolsep}{0.8mm}
\begin{tabular}{cccccc}
\toprule
\multirow{2}{*}{$\alpha$} & \multirow{2}{*}{Backbone}& \multicolumn{2}{c}{CUB-200-2011} & \multicolumn{2}{c}{Stanford Dogs} \\
\cmidrule(lr){3-4} 	\cmidrule(lr){5-6}
&&1-shot &5-shot &1-shot&5-shot\\

\midrule 
 0.1 &ResNet-12&{78.95{\scriptsize $\pm$0.83}}&{89.12{\scriptsize $\pm$0.50}}&{71.78{\scriptsize$\pm$0.91}}&{85.32{\scriptsize $\pm$0.55}}\\
 1.0&ResNet-12&{80.48{\scriptsize $\pm$0.82}}& {90.36{\scriptsize $\pm$0.46}}& \bf{73.75{\scriptsize $\pm$0.93}}&\bf{86.65{\scriptsize $\pm$0.54}}\\
5.0&ResNet-12&\bf{82.45{\scriptsize $\pm$0.79}}&\bf{92.02{\scriptsize $\pm$0.44}}& {72.65{\scriptsize $\pm$0.92}}&{86.58{\scriptsize $\pm$0.55}}\\
10.0&ResNet-12&{80.71{\scriptsize $\pm$0.82}}&{91.15{\scriptsize $\pm$0.45}}& {68.90{\scriptsize $\pm$0.92}}&{84.62{\scriptsize $\pm$0.59}}\\
\bottomrule
\end{tabular}
\label{tb:importance}
\end{table}

\subsection{Model Complexity Analysis}
The model complexity analysis is critical for model evaluation. This work presents a comparative analysis with four state-of-the-art open-source approaches to establish the superiority of the proposed approach. Two commonly used evaluation metrics, "Params.(M)" and "FLOPs(G)" are employed to analyze model complexity. Params.(M) refers to the number of parameters in the model, measured in millions. Meanwhile, FLOPs(G) denotes the number of arithmetic operations required to model feedforward, measured in billions.

As shown in Table~\ref{tb:complexity}, the proposed approach is computationally inexpensive compared to state-of-the-art FG-FSL approaches such as BSFA and AGPF. There is a significant gap in FLOPs(G) despite having similar model parameters, with only half that of the other two approaches. Moreover, the proposed approach is inferior to the traditional FSL approach CAN in terms of model complexity due to the need for a detailed exploration of the relationship between different regions. Nevertheless, the proposed approach significantly improved over CAN in classification performance (see Table~\ref{tb:comparison}). Furthermore, the proposed approach brings a negligible increase of 0.52 and 0.06 in Params.(M) and FLOPs(G), respectively. Three possible reasons account for this insignificant raise: 1) No spatial comparison in the proposed method during logit computation compared to the DN4 and BSFA. This article employs spatial descriptor aggregation followed by cost-effective channel comparison to improve computational efficiency. 2) No complex structure is introduced compared to AGPF. AGPF introduces a feature pyramid structure to capture the variations between sub-classes, which is computationally expensive, while the proposed method introduces a lightweight RHS module. 3) The proposed approach employs a dual-branch network without an increase in model complexity. During the testing phase, only the main branch is preserved.
\label{complexity}
\begin{table}[!htb]
\caption{Comparison with state-of-the-art  few-shot fine-grained visual recognition methods in model complexity. The best results are shown in bold.}
\centering
\begin{tabular}{lcll}
\toprule
\multirow{2}{*}{Method} &\multirow{2}{*}{Backbone}& \multicolumn{2}{c}{Complexity}\\
\cmidrule(lr){3-4}
~&~&Params.(M)&FLOPs(G)\\
\midrule 
DN4~\cite{li2019revisiting}&ResNet-12&12.42&67.39\\
CAN\cite{hou2019cross}&ResNet-12&8.04&12.75\\
AGPF\cite{TANG2022108792}&ResNet-12&8.77&51.53\\
BSFA\cite{zha2023boosting}&ResNet-12&8.04& 50.64\\
\midrule 
Baseline&{ResNet-12}&{8.00}&{24.80}\\
RSaD&{ResNet-12}&{8.52($\uparrow 0.52$)}&{24.86($\uparrow 0.06$)}\\
\bottomrule
\end{tabular}
\label{tb:complexity}
\end{table}

\section{Conclusion}
\label{conclusion}
This article proposed RSaD for few-shot fine-grained visual recognition. RSaD leveraged saliency-aware guidance to improve model performance while maintaining computational efficiency. Specifically, RSaD integrated multiple saliency models to produce saliency maps of superior quality. Subsequently, RSaD introduced SaG, which leveraged the distillation technique on the salient region probability distribution to exploit explicit intrinsic relationships among sub-classes. RHS was designed to highlight deep descriptors relevant to the objects of interest and squeeze them into contextual embedding to ensure transferable representation. Extensive experiments demonstrated that the proposed model achieved  comparable performance to the current state-of-the-art approach while maintaining low model complexity. 

Recently, several methods\cite{kirillov2023segment,zou2023segment,ji2023segment} have been designed to segment objects in an image all at once. These efficient and promptable methods offer more robust saliency priors than saliency detection models. Future research in this article is not limited to few-shot fine-grained classification but will focus on developing a generalized few-shot visual recognition framework based on solid segmentation models. Furthermore, there have been significant breakthroughs in semi-supervised few-shot learning methods~\cite{NEURIPS2022_5d3b57e0,10319790} in recent years. Semi-supervised learning within FS-FGVR is a promising avenue worth investigating in future work.

\bibliographystyle{IEEEtran}
\bibliography{ref}
\begin{IEEEbiography}
[{\includegraphics[width=1in,height=1.25in,clip,keepaspectratio]{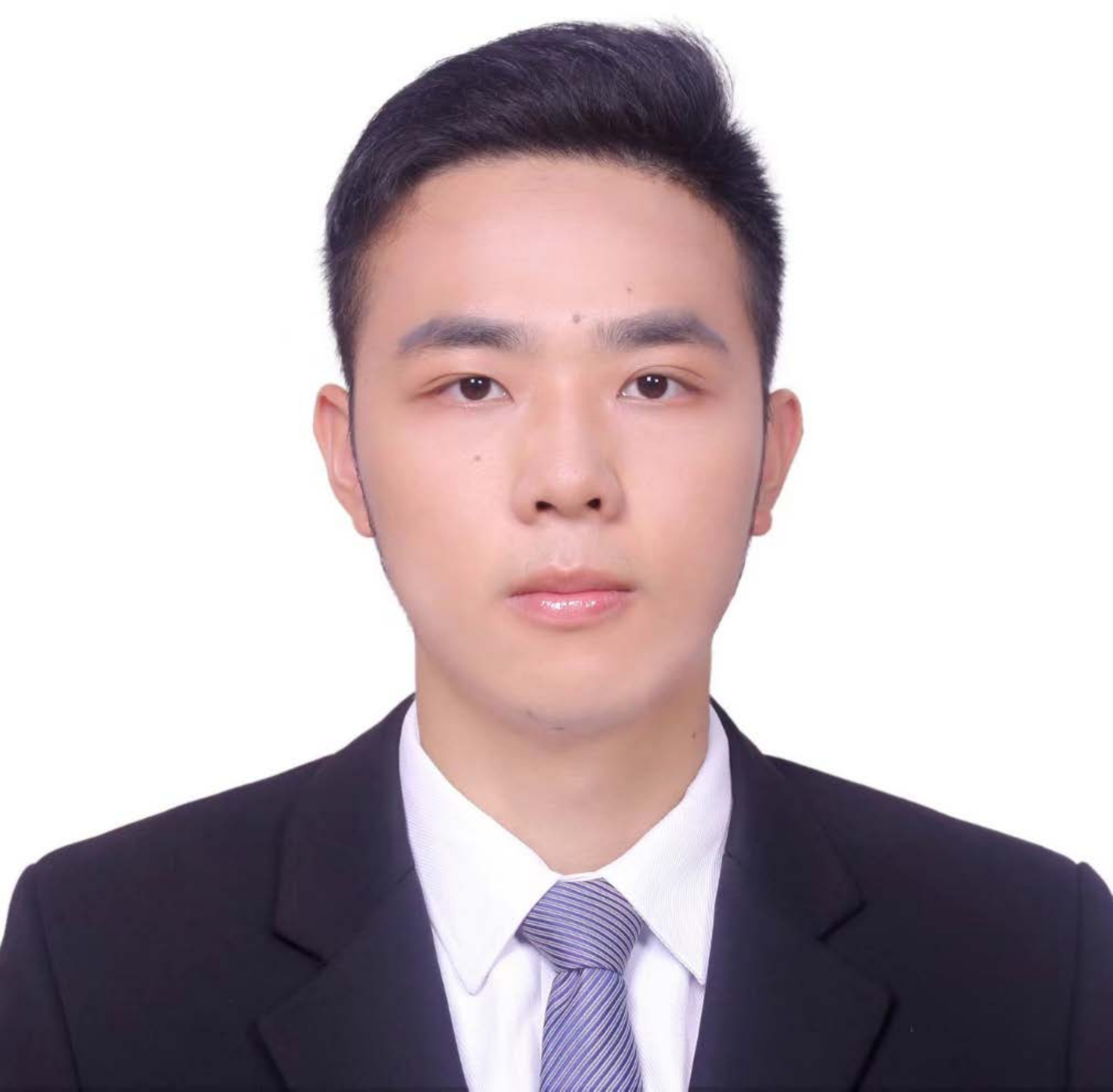}}] 
{Haiqi Liu} received the B.S. degree in computer science and technology from South China University of Technology, Guangzhou, China, in 2020. He is currently pursuing the Ph.D. degree in computer science and technology with the School of Computer Science and Engineering. 

His research interests mainly include Few shot Learning, Image Recognition and Affective Computing.
\end{IEEEbiography}
\begin{IEEEbiography}
[{\includegraphics[width=1in,height=1.25in,clip,keepaspectratio]{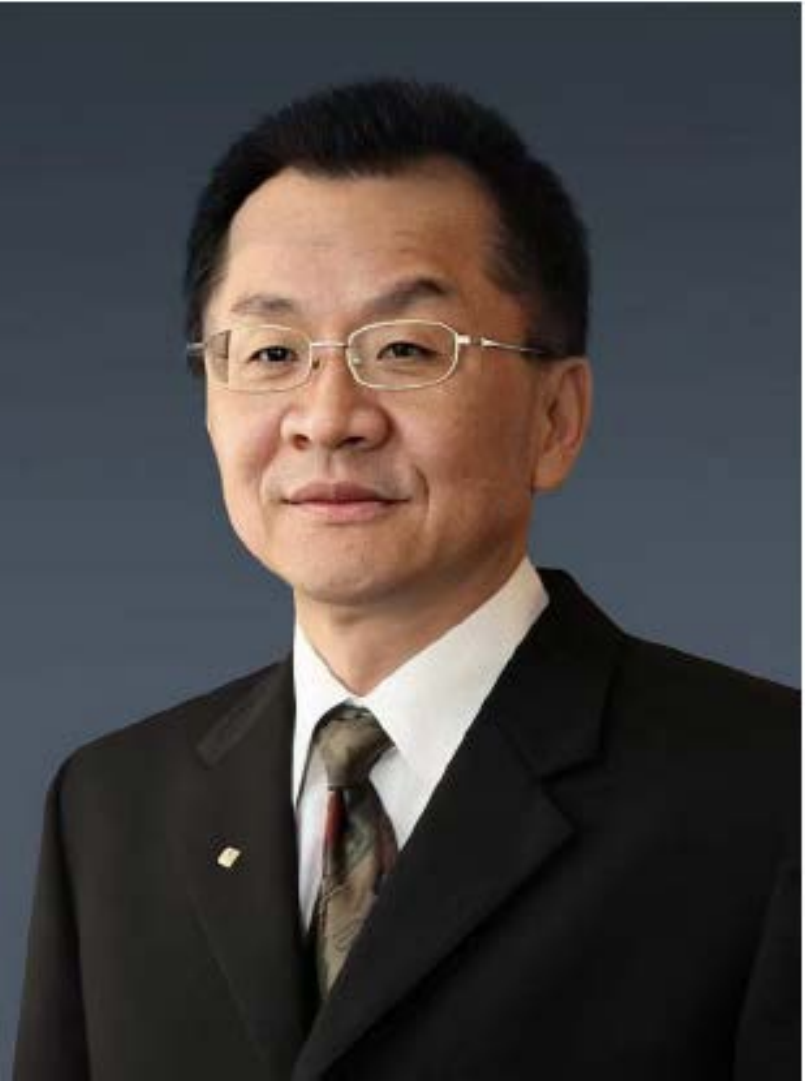}}] 
{C. L. Philip Chen}(S’88–M’88–SM’94–F’07) received the M.S. degree from the University of Michigan at Ann Arbor, Ann Arbor, MI, USA, in 1985 and the Ph.D. degree from the Purdue University in 1988, all in electrical and computer science.

He is the Chair Professor and Dean of the College of Computer Science and Engineering, South China University of Technology. He is the former Dean of the Faculty of Science and Technology. He is a Fellow of IEEE, AAAS, IAPR, CAA, and HKIE; a member of Academia Europaea (AE) and European Academy of Sciences and Arts (EASA). He received IEEE Norbert Wiener Award in 2018 for his contribution in systems and cybernetics, and machine learnings. He is also a highly cited researcher by Clarivate Analytics in 2018, 2019, 2020, 2021 and 2022.

He was the Editor-in-Chief of the IEEE Transactions on Cybernetics (2020-2021) after he completed his term as the Editor-in-Chief of the IEEE Transactions on Systems, Man, and Cybernetics: Systems (2014-2019), followed by serving as the IEEE Systems, Man, and Cybernetics Society President from 2012 to 2013. Currently, he serves as an deputy director of CAAI Transactions on AI, an Associate Editor of the IEEE Transactions on AI, IEEE Trans on SMC: Systems, and IEEE Transactions on Fuzzy Systems, an Associate Editor of China Sciences: Information Sciences. He received Macau FDCT Natural Science Award three times and a First-rank Guangdong Province Scientific and Technology Advancement Award in 2019. His current research interests include cybernetics, computational intelligence, and systems.
\end{IEEEbiography}
\begin{IEEEbiography}[{\includegraphics[width=1in,height=1.25in, clip,keepaspectratio]{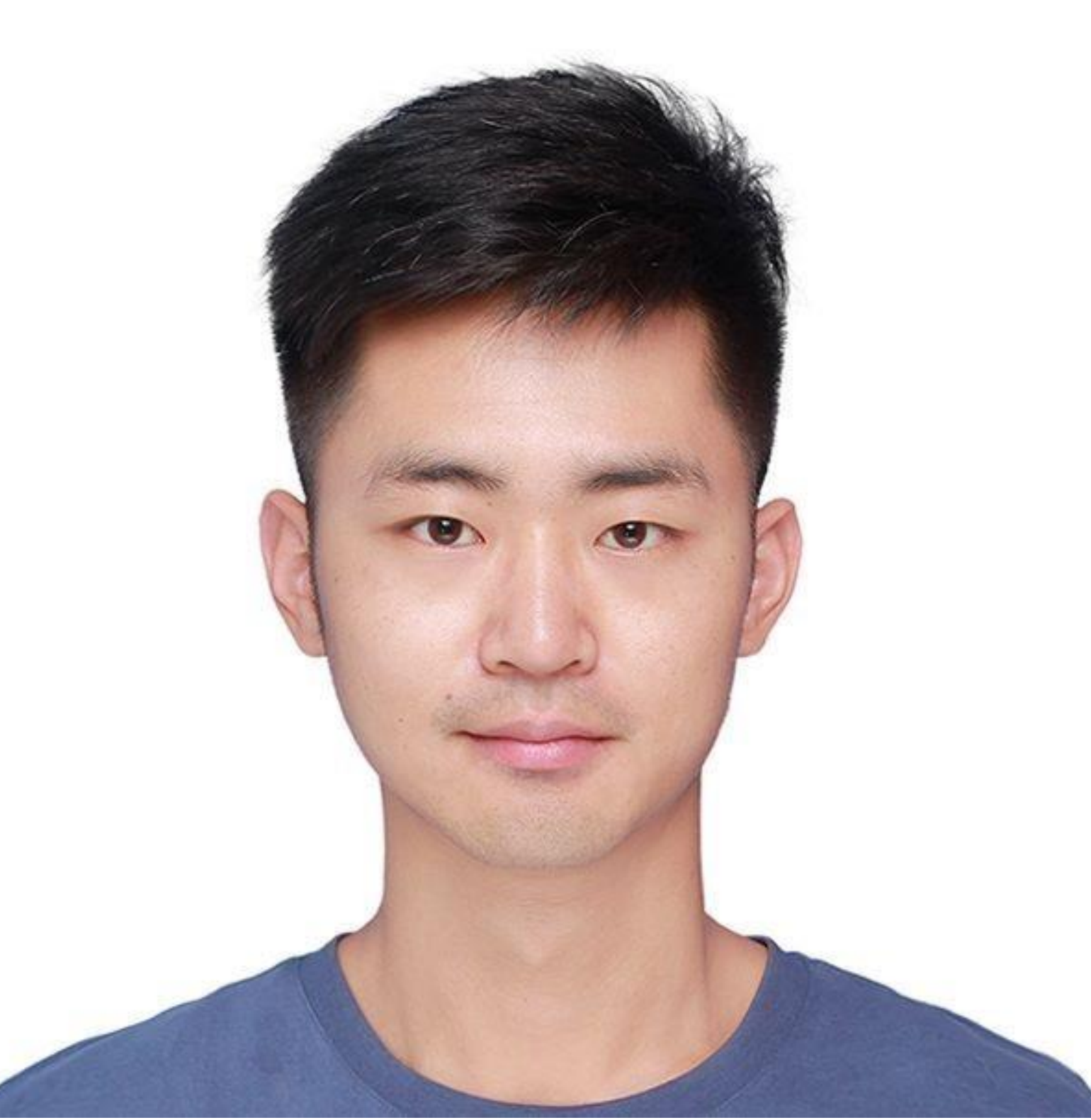}}]{Xinrong Gong} received the B.S. degree in the School of Electronic and Optical Engineering from the Nanjing University of Science and Technology, Nanjing, China, in 2018. He is pursuing a Ph.D. at the School of Computer Science and Engineering at the South China University of Technology, in Guangzhou, China. 

His research interests include Machine Learning, Broad Learning Systems, and Multimodal Emotion Recognition.
\end{IEEEbiography}
\begin{IEEEbiography}
[{\includegraphics[width=1in,height=1.25in,clip,keepaspectratio]{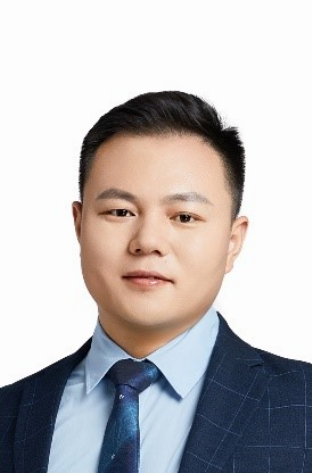}}] 
{Tong Zhang} (M'16-SM’24) received the B.S. degree in software engineering from Sun Yat-sen University, at Guangzhou, China, in 2009, and the M.S. degree in applied mathematics from University of Macau, at Macau, China, in 2011, and the Ph.D. degree in software engineering from the University of Macau, at Macau, China in 2016. Dr. Zhang currently is a professor and Associate Dean of the School of Computer Science and Engineering, South China University of Technology, China. 

His research interests include affective computing, evolutionary computation, neural network, and other machine learning techniques and their applications. Prof. Zhang is the Associate Editor of the IEEE Transactions on Affective Computing, IEEE Transactions on Computational Social Systems, and Journal of Intelligent Manufacturing. He has been working in publication matters for many IEEE conferences. 
\end{IEEEbiography}

\vfill

\end{document}